\documentclass[11pt]{cambrian}
\usepackage[utf8]{inputenc}
\DeclareUnicodeCharacter{207A}{\textsuperscript{+}}
\usepackage{graphicx}
\usepackage{multirow}    
\usepackage{booktabs}    
\usepackage{float}       
\usepackage{amsmath}      
\usepackage{subcaption}   
\usepackage{tcolorbox}    
\tcbset{floatplacement=t}

\usepackage{xspace}

\usepackage{subcaption}   
\def\mycolor{\cellcolor[HTML]{CFEFFF}}
\newcommand{\prompt}{Reasoning Instruction\xspace}
\newcommand{\ourmethod}{ViGaL\xspace}
\newcommand{\tablestyle}[2]{\setlength{\tabcolsep}{#1}\renewcommand{\arraystretch}{#2}\centering\small}

\makeatletter\renewcommand\paragraph{\@startsection{paragraph}{4}{\z@}
{.2em \@plus1ex \@minus.2ex}{-.5em}{\normalfont\normalfont\bfseries}}
\makeatother

\newcolumntype{x}[1]{>{\centering\arraybackslash}p{#1pt}}
\newcolumntype{y}[1]{>{\raggedright\arraybackslash}p{#1pt}}
\newcolumntype{z}[1]{>{\raggedleft\arraybackslash}p{#1pt}}
\usepackage[utf8]{inputenc}

\usepackage[numbers,sort&compress]{natbib}
\usepackage{colortbl}

\usepackage[utf8]{inputenc} 
\usepackage[T1]{fontenc}    
\usepackage{hyperref}
\usepackage{url}            
\usepackage{booktabs}       
\usepackage{amsfonts}       
\usepackage{nicefrac}       
\usepackage{microtype}      
\usepackage{xcolor}
\definecolor{bluelink}{RGB}{0,113,188}
\definecolor{greenlink}{RGB}{0,188,113}
\hypersetup{
    colorlinks=true,%
    citecolor=green!93!black,%
    filecolor=redlink,%
    linkcolor=black,%
    urlcolor=bluelink
}
\usepackage{tabularx}
\usepackage{tcolorbox}
\usepackage{amsmath}
\usepackage{multirow}
\usepackage{array}
\usepackage{caption}
\usepackage{wrapfig}
\usepackage{enumitem}
\usepackage{tikz}
\usepackage{lipsum}
\usepackage{subcaption}
\usepackage{multirow} 
\usepackage{adjustbox}

\usepackage{amssymb}
\usepackage{unicode}  
\usepackage[capitalize]{cleveref} 

\usepackage{pgf}
\usepackage{colortbl}

\usepackage{tipa}

\usepackage{rotating}
\usepackage[abs]{overpic}
\usepackage{makecell}
\usepackage{longtable}

\usepackage{tocloft}  

\usepackage[english]{babel}
\usepackage{csquotes}
\usepackage{listings}
\definecolor{codekeyword}{rgb}{0.0, 0.0, 0.5}   
\definecolor{codecomment}{rgb}{0.0, 0.5, 0.0}   
\definecolor{codestring}{rgb}{0.56, 0.0, 1.0}   

\lstdefinestyle{pythonstyle}{
    language=Python,                          
    basicstyle=\ttfamily\small,               
    keywordstyle=\color{codekeyword}\bfseries,
    commentstyle=\color{codecomment}\itshape, 
    stringstyle=\color{codestring},           
    showstringspaces=false,                   
    breaklines=true,                          
    tabsize=4,                                
    numbers=none,                             
    frame=none,                               
    backgroundcolor=\color{white},            
    captionpos=b,                             
    morekeywords={self, __init__, __name__, __main__}, 
}

\usepackage{fontawesome}
\usepackage{xspace}

\newcommand{\github}{\raisebox{-1.5pt}{\includegraphics[height=1.05em]{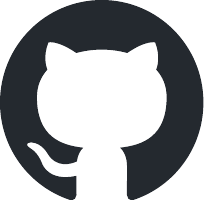}}\xspace}
\newcommand{\worldwideweb}{\raisebox{-1.5pt}{\includegraphics[height=1.05em]{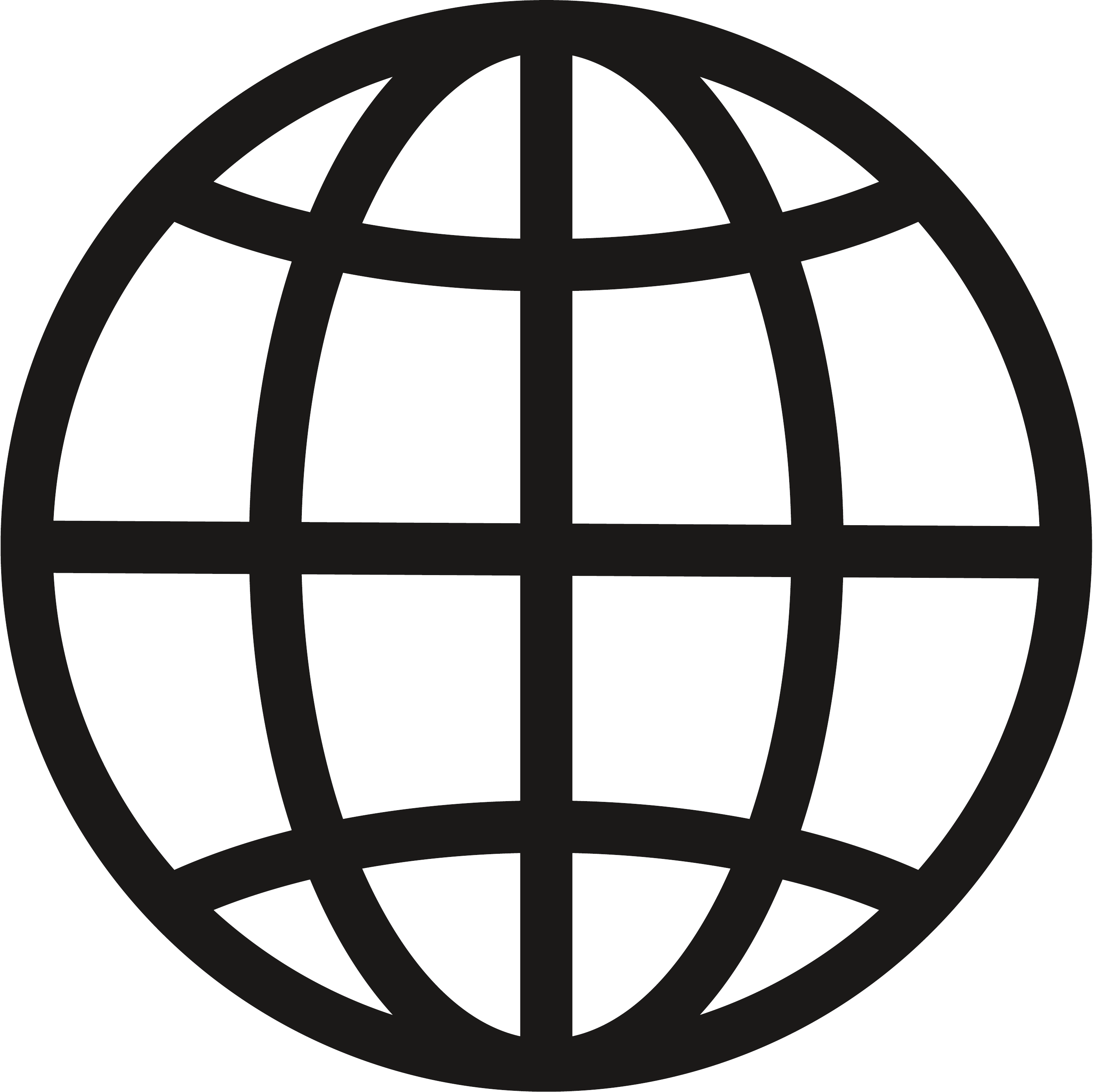}}\xspace}

\title{\centering Play to Generalize: \\ Learning to Reason Through Game Play}

\author{%
Yunfei~Xie\textsuperscript{1}, 
Yinsong~Ma\textsuperscript{2}, 
Shiyi~Lan\textsuperscript{3}, 
Alan~Yuille\textsuperscript{2}, 
Junfei~Xiao\textsuperscript{2}\footnote{Project Lead; \textsuperscript{\textdagger}Corresponding Author}, 
Chen~Wei\textsuperscript{1\textdagger}\\
\textsuperscript{1}Rice University,\quad \textsuperscript{2}Johns Hopkins University,\quad \textsuperscript{3}NVIDIA\\

  \vspace{2pt}
  {\small
    \begin{tabular}{@{}rll@{}}
      \worldwideweb & \textbf{Website}           & \url{https://yunfeixie233.github.io/ViGaL}\\
      \github       & \textbf{Code \& Model \& Data} & \url{https://github.com/yunfeixie233/ViGaL}
    \end{tabular}%
  }
}

\def\mycolor{\cellcolor[HTML]{CFEFFF}}
\newcolumntype{x}[1]{>{\centering\arraybackslash}p{#1pt}}
\newcolumntype{y}[1]{>{\raggedright\arraybackslash}p{#1pt}}
\newcolumntype{z}[1]{>{\raggedleft\arraybackslash}p{#1pt}}

\begin{document}

\begin{abstract}

Developing reasoning capabilities in multimodal large language models (MLLMs) remains challenging. Motivated by literature suggesting that gameplay promotes transferable reasoning skills, we propose a novel post-training method, Visual Game Learning (ViGaL), where MLLMs develop generalizable reasoning skills through playing arcade-like games. Specifically, we show that training a 7B-parameter MLLM via reinforcement learning (RL) on simple games like Snake significantly enhances the downstream performance on multimodal math benchmarks like MathVista, on multi-discipline questions like MMMU and on 3D spatial reasoning benchmarks like VSI-Bench, without seeing any worked solutions, equations, or diagrams during RL.
Remarkably, our model outperforms specialist models post-trained on benchmark-oriented multimodal reasoning data, while preserving the model’s performance on general visual benchmarks, a challenge where specialist models often fall short.
Our findings suggest that multimodal reasoning can emerge from gameplay, pointing to a promising strategy of designing surrogate tasks for RL post-training.
\end{abstract}
 
\maketitle

\begin{figure}[H]
  \centering

  \begin{subfigure}[b]{0.57\textwidth}
    \centering
    \includegraphics[width=\linewidth]{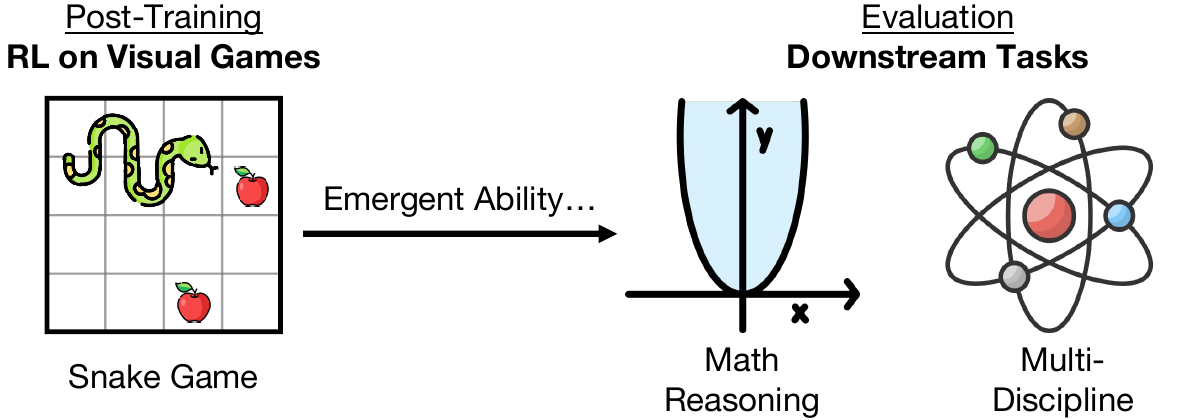}
    \vspace{-10pt}
    \label{fig:teaser_fig}
  \end{subfigure}%
  \hfill
    \begin{subfigure}[b]{0.38\textwidth}
    \centering
    \includegraphics[width=\linewidth]{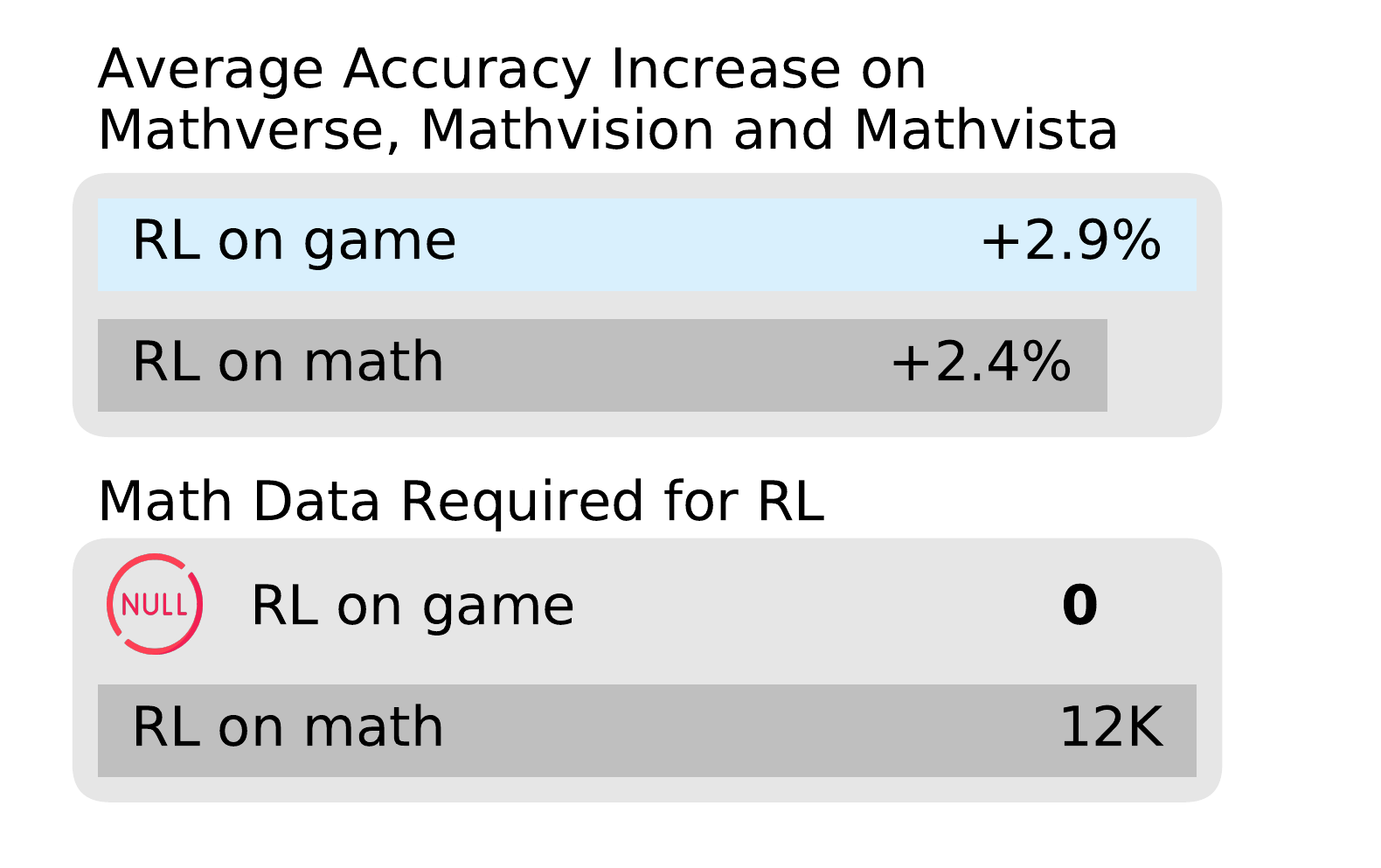}
    \vspace{-20pt}
    \label{fig:teaser_bar}
  \end{subfigure}%
\caption{\textbf{Overview of \ourmethod.} \textit{Left}:
We propose a novel post-training method where MLLMs are finetuned via RL to play arcade-style games such as Snake~\citep{snake_bench_2025}. We demonstrate that gameplay post-training enables MLLMs to achieve {out-of-domain} generalization, enhancing their performance on downstream multimodal reasoning tasks requiring math, spatial and multi-discipline reasoning, without using math or multi-displine data during RL.
\textit{Right}:
Our \ourmethod (RL on game) achieves higher average accuracy increase than MM-Eureka~\citep{meng2025mm} (RL on math) across three multimodal math benchmarks. This is notable because MM-Eureka trains on large-scale, curated math datasets, while \ourmethod only uses game data. Details are in Tab~\ref{tab:math_generalization}.}
\label{fig:teaser}
\end{figure}
 
\section{Introduction}
\vspace{5pt}
\label{sec:intro}

Games, beyond their entertainment value, provide rich and diverse structured environments for developing and studying general reasoning and problem-solving abilities. Humans from early childhood acquire foundational cognitive skills through diverse game-like activities such as arranging objects, navigating spaces, and manipulating tools. These experiences foster essential building blocks of abstract thinking, including pattern recognition, spactial reasoning, and causal inference~\citep{brandle2021using,bertram2020digital}.
In cognitive science, games are used as experimental platforms to reveal the inductive biases of the human mind~\citep{allen2024using,alhasoun2021probabilistic}, such as planning depth in the game Four-in-a-Row~\citep{van2023expertise}, or the cognitive basis of tool use through the game Virtual Tools~\citep{allen2020rapid}.

AI agents, too, have benefited from games resembling aspects of human play. These environments encourage exploration, robustness to sparse rewards, and learning from multimodal inputs. For example, emergent tool use has been observed in agents trained via hide-and-seek~\citep{baker2019emergent}, and Atari gameplay has been incorporated into training generalist agents~\citep{reed2022generalist}. By learning in these environments, AI systems develop robust and transferable reasoning capabilities.

Recent work has shown that post-training with Reinforcement Learning (RL) can unlock reasoning behaviors from their base models~\citep{deepseekai2025,openai2024o1}. These RL-trained models can ``think before they speak'', generating internal chain-of-thought traces before outputting a final answer.
More importantly, growing evidence suggests that RL often generalizes more robustly to out-of-distribution samples than supervised fine-tuning (SFT).
For example, models trained with RL on mathematics transfer their reasoning skills to physics~\citep{meng2025mm}, and navigation agents adapt to novel environments beyond their training domains~\citep{chu2025sft}. Motivated by these findings, we ask: \textit{since games already serve as a natural medium through which humans acquire reasoning strategies, can post-training multimodal LLMs on gameplay similarly enhances their ability to reason across diverse tasks?}


The results are striking (Fig.~\ref{fig:teaser}). We show that post-training a 7B-parameter multimodal LLM, Qwen2.5-VL-7B~\citep{Qwen2.5-VL}, to play simple arcade-style games like Snake~\citep{snake_bench_2025} yields two surprising outcomes:
(1) the model generalizes to previously unseen games (Sec.~\ref{sec:game_generalization}); and
(2) it exhibits strong reasoning abilities on multimodal math benchmarks like MathVista~\citep{lu2024mathvistaevaluatingmathematicalreasoning}, and multi-domain QA like MMMU~\citep{yue2024mmmu}.
Despite never observing worked solutions, equations, or diagrams during RL post-training, the model achieves competitive results not only against large-scale industrial systems like GPT-4o~\citep{hurst2024gpt}, but also against specialist models post-trained on math datasets (Tabs.~\ref{tab:math_generalization} and \ref{tab:reason_generalization}). Furthermore, it improves on reasoning benchmarks without degrading general visual understanding, a common limitation of domain-specialist training (Tab.~\ref{tab:vision_generalization}).
Overall, gameplay emerges as an effective surrogate task for incentivizing reasoning in multimodal LLMs.

Why does it work? Our ablation studies suggest that reasoning skills incentivized by gameplay can be helpful to other multimodal reasoning tasks. For example, Snake, a game set on a \textit{2D grid} where the player maneuvers the ``snake'' to avoid collisions and collect apples, significantly improves performance on math problems involving \textit{2D coordinates}. In contrast, Rotation, a puzzle requiring recognition of 3D \textit{object rotation angles}, more strongly boosts performance on \textit{geometry questions involving angles and lengths} (Fig.~\ref{fig:acc_diff}). Furthermore, jointly training on both games yields consistently stronger results on downstream benchmarks than training on either game alone, suggesting the \textbf{compositionality} of the acquired skills (Tab.~\ref{tab:math_generalization}).

\begin{figure}[t]
    \centering
    \includegraphics[width=0.99\linewidth]{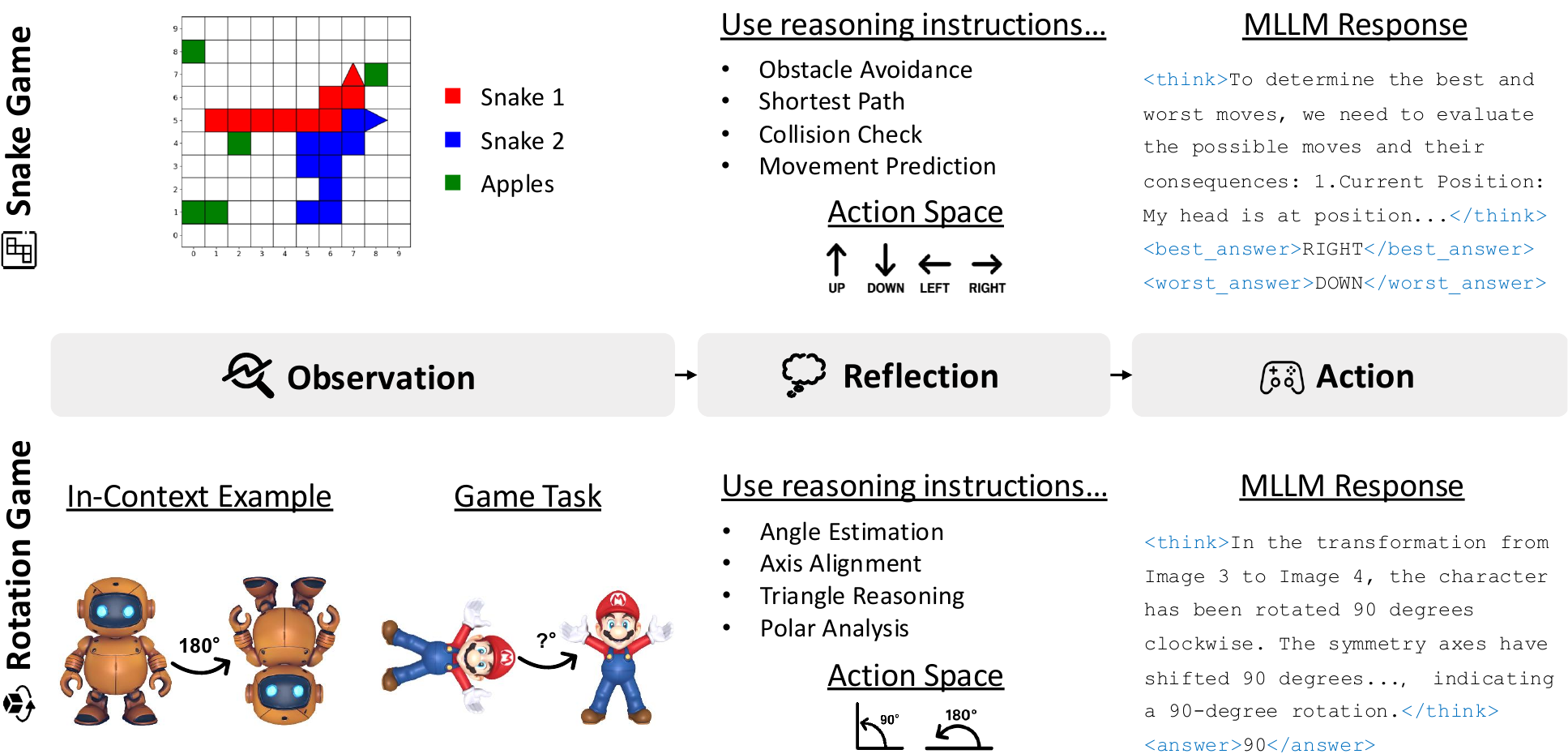}
    \caption{\textbf{Post-training MLLMs to reason through RL with games.} We propose post-training MLLMs via RL by playing visual games. We demonstrate this with two games: the classic arcade game Snake~\citep{snake_bench_2025}, and Rotation, a self-designed task to investigate spatial reasoning. In each game, the model receives multimodal inputs and follows reasoning instructions, \textit{e.g.}, path planning in Snake, angle estimation in Rotation. It reflects to choose an action, outputs its chain-of-thoughts and decision, \textit{e.g.}, best/worst move or predicted angle, and receives a reward. Through gameplay, the model obtains reasoning abilities that transfer to downstream multimodal reasoning tasks such as math and multi-discipline question answering.
    }
    \label{fig:game_pipeline}
\end{figure}

All these results point to a new post-training strategy: rather than relying solely on domain-specific datasets, we can design scalable and controllable \textbf{surrogate tasks for post-training}, such as games, that unlock reasoning behaviors transferable to downstream applications. Synthetic game environments offer structured, rule-based rewards and fine-grained controllability, while also scaling far more easily than human-annotated data. This promising paradigm of post-training with surrogate tasks reminisces self-supervised pre-training in vision and language~\citep{he2020momentum,doersch2015unsupervised,radford2018improving}, where carefully designed pretext tasks produce broad generalization.




\section{Reinforcement Learning on Visual Games}
\vspace{-5pt}
\label{sec:RL_game}
We introduce \ourmethod{}, a novel post-training paradigm designed to enhance generalization capabilities. 

\vspace{-5pt}
\subsection{Game Environment}
\vspace{-5pt}
\label{sec:game_env}
As show in Fig.~\ref{fig:game_pipeline}, under our ViGaL paradigm, the model is trained in a game environment where it receives states from game environment, outputs next actions, and obtains rewards as feedback from the environment. Formally, each task, given an instruction $I$, can be formulated as a partially observable Markov decision process (POMDP): $(\mathcal{S}, \mathcal{A}, \mathcal{O}, T, R, \Omega)$, where $\mathcal{S}$ is the set of possible environment states, $\mathcal{O}$ is the set of observations available to the model, and $\mathcal{A}$ represents actions model can do in this game environment.
$T: \mathcal{S} \times \mathcal{A} \to \mathcal{S}$ is the state transition function, while $R$ is a binary reward from the environment representing the correctness of action. Due to partial observability, the agent perceives only observations $o = \Omega(s)$.

\paragraph{Snake and Rotation Games.} 
We design two complementary games, Snake and Rotation, to study the proposed paradigm (Fig.~\ref{fig:game_pipeline}), each focusing on different MLLM capabilities. The Snake game, inspired by prior work showing that competition can enhance reasoning in MLLMs~\citep{du2023improving}, emphasizes strategic decision-making. We set up a dual-snake game based on SnakeBench~\citep{snake_bench_2025}, where each model independently controls one snake. The objective is to reach apples, score points, and outcompete the opponent. At time $t$, the environment state $s^t$ includes the coordinates of both snakes $(x_{s_i}^t, y_{s_i}^t)$ for $i \in \{1, 2\}$, the apple location $(x_a^t, y_a^t)$, and the previous actions $A_i^{t-1}$. All elements are placed on a $10 \times 10$ board. Each snake then selects its next action $A_i^t \in \{\texttt{up}, \texttt{down}, \texttt{left}, \texttt{right}\}$. A snake dies if it collides with itself, the other snake, or the board boundary; the survivor wins, or in the case of simultaneous death, the higher score decides. Unlike SnakeBench, which uses only text to represent states, we provide both images of the game board and textual descriptions as observations $o^t = \Omega(s^t)$ for richer input. The Rotation game, inspired by rotation-angle prediction as a pre-text task in self-supervised learning~\citep{gidaris2018unsupervised}, evaluates visual perception and spatial reasoning. The model is presented with two views of the same 3D object: an initial view $I_{\textrm{init}}$ and a rotated view $I_{\textrm{rot}}$, obtained by rotating the object $90^{\circ}$ or $180^{\circ}$ around the $z$-axis (pointing toward the viewer). The task is to identify which rotation angle transforms $I_{\textrm{init}}$ into $I_{\textrm{rot}}$. To guide reasoning, we include an in-context example with a known rotation. As in the Snake game, observations combine images and text. Together, these two games allow systematic exploration of reasoning and perception, two fundamental aspects of MLLM abilities.

\subsection{Rule-Based Reinforcement Learning}
\label{sec:RL_algo}
We apply rule-based RL to directly post-train MLLMs for visual games, without relying on supervised learning as a warm up. The algorithm is described as follows:

\paragraph{Reward design.}

We use a simple rule-based reward function to avoid reward hacking~\citep{gao2022scalinglawsrewardmodel} and help the model learn how to play the games effectively. This reward function has two components: an accuracy reward and a format reward. The total reward $r$ is computed as the sum of an accuracy reward and a format reward $r = r_\text{accuracy} + r_\text{format}$. The accuracy reward \( r_\text{accuracy} \) is 1 if the answer is correct, and 0 otherwise. Details of reward for each game are in Appendix Sec.~\ref{sec:detail_reward}.

\paragraph{Advantage estimation and policy update.} We employ REINFORCE Leave-One-Out (RLOO) algorithm~\citep{RLOOKoolHW19a, RLOOAhmadianCGFKPUH24} in our RL training phase.
Following Group Policy Gradient~\citep{chu2025gpg}, we omit KL divergence regularization. Without KL constraints limiting policy changes, the model explores the solution space more freely, potentially discovering better reasoning strategies. This enables more flexible adaptation during RL training.


\subsection{Implementation and Evaluation on Games}

\begin{table}[t]
\centering
\begin{minipage}{0.25\textwidth}
    \centering
    \tablestyle{1pt}{1.05}
    \scalebox{0.85}{
    \begin{tabular}{lc}
        \textbf{Model} & \textbf{Wins (/10)} \\
        \midrule
        \mycolor \ourmethod vs. & \mycolor \\
        \hspace{0.8em} Qwen2.5-VL-7B &  9 \\
        \hspace{0.8em} Qwen2.5-VL-72B &  7 \\
        \hspace{0.8em} Llama-4-Maverick &  7 \\ 
        \hspace{0.8em} Gemini-2.5-Pro & 8 \\
        \hspace{0.8em} Claude-3.7-Sonnet &  6 \\ 
        \hspace{0.8em} GPT-4o & 8 \\
        \hspace{0.8em} o4-mini & 6 \\
    \end{tabular}
    }
    \subcaption{\centering Snake game.}
    \label{tab:snakegame}
\end{minipage}
\hfill
\begin{minipage}{0.25\textwidth}
    \centering
    \tablestyle{3pt}{1.05}
    \scalebox{0.85}{
    \begin{tabular}{lc}
        \textbf{Model} & \textbf{Acc. (\%)} \\
        \midrule
        \mycolor \ourmethod & \mycolor \textbf{71.9} \\      
        \hspace{0.8em} Qwen2.5-VL-7B & 47.4 \\
        \hspace{0.8em} Qwen2.5-VL-72B & 52.1 \\
        \hspace{0.8em} Llama-4-Maverick & 66.2 \\
        \hspace{0.8em} Gemini-2.5-Pro & 51.0 \\
        \hspace{0.8em} Claude-3.7-Sonnet & 65.6 \\
        \hspace{0.8em} GPT-4o & 61.5 \\
        \hspace{0.8em} o4-mini & 70.8 \\
    \end{tabular}
    }
    \subcaption{\centering Rotation game.}
    \label{tab:rotationgame}
\end{minipage}
\hfill
\begin{minipage}{0.48\textwidth}
\centering
\tablestyle{3pt}{1.0}
\scalebox{0.85}{
\begin{tabular}{lcc}
\textbf{Game} & \textbf{\ourmethod} & \textbf{Qwen2.5-VL-7B} \\
\midrule
Space Invaders & 280.0 & 85.0 \\
Ms. Pacman & 1370.0 & 670.0 \\
Seaquest & 80.0 & 60.0 \\
Alien & 540.0 & 450.0 \\
Frogger & 7.0 & 5.0 \\
Breakout & 0.0 & 9.0 \\
Pong & -26.0 & -26.0 \\
\midrule
Cumulative Reward & \textbf{2251.0} & 1253.0 \\
\end{tabular}
}
\subcaption{\centering Atari game.}
\label{tab:atarigame}
\end{minipage}
\vspace{-5pt}
\caption{\textbf{Game Performance.} (a) \ourmethod gets high win rates (6-9 wins out of 10 matches) on Snake playing against advanced proprietary models. (b) \ourmethod shows best performance on Rotation. (c) \ourmethod trained on Snake and Rotation shows zero-shot generalization to unseen Atari games, achieving a nearly \textit{doubled} cumulative reward compared to its base model (Qwen2.5-VL-7B).}
\vspace{-10pt}
\label{tab:game_performance}
\end{table}
\label{sec:game_perf}
\paragraph{Implementation details.}
We employ Qwen2.5-VL-7B-Instruct~\citep{Qwen2.5-VL} as our base model. We follow DeepSeek-R1~\citep{deepseekai2025}, using a combination of rule-based format rewards and accuracy rewards, with RLOO~\citep{RLOOKoolHW19a, RLOOAhmadianCGFKPUH24} as the core RL algorithm. We implement our training within a multimodal input RL framework based on OpenRLHF~\citep{hu2024openrlhf}.
For hyperparameters, we adopt the default settings from MM-Eureka~\citep{meng2025mm}, including a global batch size of 128, a rollout batch size of 128, a rollout temperature of 1.0, and a learning rate of $1e^{-6}$. Training uses 6 A100-80G GPUs.

\paragraph{Game training data.}
\label{sec:data}
We build game environments to collect training data for our experiments. For Snake, we leverage SnakeBench~\citep{snake_bench_2025} as our data engine. For Rotation, we utilize Hunyuan3D~\citep{hunyuan3d22025tencent}, which generates 3D meshes from images or text instructions. We render each mesh into 2D images from different orientations, creating image pairs with associated rotation angles as ground truth labels for RL training. Our comprehensive data generation pipeline enables producing training samples at any desired scale with fully customized settings. For experiments, we synthesize 36K samples per game, sufficient for convergence. Details are in Appendix Sec.~\ref{sec:data_appendix}.

\paragraph{Competing with leading models on Snake and Rotation.} To evaluate the game capabilities of \ourmethod{} models, we initialize environments in diverse states unseen during training. For Snake (Tab.~\ref{tab:snakegame}), we randomly initialize games 10 times with two models competing directly, measuring win counts. For Rotation (Tab.~\ref{tab:rotationgame}), we measure rotation angle prediction accuracy on comprehensive validation sets with 3D object meshes unseen during training. Our 7B-parameter model consistently outperforms proprietary models in both games. Results confirm that RL effectively unlocks small 7B models' ability to excel in visual games requiring environmental understanding, reasoning, planning, and interactive decision-making.

\paragraph{Out-of-distribution generalization to Atari games.}
\label{sec:game_generalization}
We then test \ourmethod on Atari-GPT~\citep{waytowich2024atari}, a benchmark for evaluating MLLMs as decision-making agents in Atari video games such as in Fig.~\ref{fig:atarigame}. The benchmark consists of seven different Atari games, with detailed settings in Appendix Sec.~\ref{sec:atarigame}. We follow most settings and prompts from Atari-GPT, with a small modification to ensure format correctness for all models. Following Atari-GPT~\citep{waytowich2024atari}, we report cumulative reward over 1K steps as the evaluation metric, where higher rewards indicate better performance.
As shown in Tab.~\ref{tab:atarigame}, \ourmethod shows significant cumulative reward improvement on Atari games despite being trained only on Snake and Rotation games. This is particularly notable because Atari games differ substantially from our training games in both visual appearance and gameplay strategies. These results suggest that our rule-based RL training approach enables strong generalization to previously unseen game environments.

\newcommand{\tightpm}[1]{\raisebox{0pt}[0pt][0pt]{\tiny $\pm$#1}}
\begin{table}[t]
    \centering
    \tablestyle{2pt}{1.2}
    \resizebox{\linewidth}{!}{
    \begin{tabular}{l|>{\columncolor{gray!8}}c|
                    >{\columncolor{gray!8}}cccc|
                    >{\columncolor{gray!8}}ccc}
        & \multirow{2}{*}{\textbf{\begin{tabular}{@{}c@{}}\\Avg.\end{tabular}}}
        & \multicolumn{4}{c|}{\textbf{Math}}
        & \multicolumn{3}{c}{\textbf{Geometry}}\\
        \multirow{-2}{*}{\textbf{Model}}
        & \textbf{Avg.}
        & \textbf{Avg.} & \textbf{MathVista} & \textbf{MathVerse} & \textbf{MathVision}
        & \textbf{Avg.} & \textbf{GeoMath} & \textbf{Geo3K}\\
        \midrule
        \multicolumn{9}{@{}c}{Proprietary Model}\\
        \midrule
        GPT-4o~\citep{hurst2024gpt} & 47.5 & 47.3 & 61.4 & 50.2 & 30.4 & 46.8 & 50.2 & 43.5\\
        Gemini-2.0-Flash~\citep{team2023gemini} & 55.4 & 56.4 & 73.4 & 54.6 & 41.3 & 54.4 & 55.3 & 53.5\\
        \midrule \midrule
        \multicolumn{9}{@{}c}{Multimodal Reasoning Model Post-Trained on Qwen2.5-VL-7B~\citep{Qwen2.5-VL}}\\
        \midrule
        \textit{Base Model (Qwen2.5-VL-7B)}
            & 46.3 & 47.7 & 68.0 & 49.0 & 26.0 & 44.8 & 44.0 & 45.6\\
        \hline
        R1-Onevision-7B~\citep{yang2025r1} & 40.9 & \textcolor{gray}{46.8} & \textcolor{gray}{64.1} & \textcolor{gray}{46.4} & \textcolor{gray}{\textbf{29.9}}
            & 35.0 & 45.4 & 24.5\\
        R1-VL-7B~\citep{chen2025r1v} & 40.9 & \textcolor{gray}{42.7} & \textcolor{gray}{63.5} & \textcolor{gray}{40.0} & \textcolor{gray}{24.7}
            & \textcolor{gray}{39.0} & \textcolor{gray}{42.0} & \textcolor{gray}{36.1}\\
        MM-Eureka-Qwen-7B~\citep{meng2025mm} & 39.3 & \textcolor{gray}{50.1} & \textcolor{gray}{73.0} & \textcolor{gray}{50.3} & \textcolor{gray}{26.9}
            & 28.4 & 53.1 & 3.8\\
        Reason-RFT-Zero-7B~\citep{tan2025reason} & 46.5 & 38.1 & 60.7 & 35.3 & 18.3
            & \textcolor{gray}{54.9} & \textcolor{gray}{55.0} & \textcolor{gray}{54.8}\\
        VLAA-Thinker-7B~\citep{chen2025sftrlearlyinvestigation}
            & 51.3 & \textcolor{gray}{48.7} & \textcolor{gray}{68.0} & \textcolor{gray}{51.7} & \textcolor{gray}{26.4}
            & \textcolor{gray}{53.9} & \textcolor{gray}{51.1} & \textcolor{gray}{56.6}\\
        OpenVLThinker-7B~\citep{deng2025openvlthinker}
            & 52.1 & \textcolor{gray}{47.8} & \textcolor{gray}{70.2} & \textcolor{gray}{47.9} & \textcolor{gray}{25.3}
            & \textcolor{gray}{56.4} & \textcolor{gray}{49.2} & \textcolor{gray}{63.5}\\
        \hline
        \ourmethod{} Snake & 51.6 & 49.4 & 70.7 & 51.1 & 26.5 & 55.0 & 49.9 & 60.0\\
        \ourmethod{} Rotation                  & 52.8 & 49.3 & 71.2 & 50.4 & 26.3 & \textbf{57.9} & 51.7 & 64.1\\
        
        \ourmethod{} Snake + Rotation & \textbf{53.9} & \textbf{50.6} & 71.9 & 52.4 & 27.5 & 57.1 & 51.0 & 63.3 \\
         \noalign{\vskip-0.8ex} 
        & \tightpm{0.3} & \tightpm{0.3} & \tightpm{0.4} & \tightpm{0.2} & \tightpm{0.3} & \tightpm{0.5} & \tightpm{0.3} & \tightpm{0.4} \\

    \end{tabular}}
\vspace{-5pt}
\caption{\textbf{Results on multimodal mathematical benchmarks.} We compare to other multimodal reasoning models. 
Results post-trained on the same subject as the evaluation are \textcolor{gray}{de-emphasized}, while our \ourmethod{} models only use games for post-training. \textbf{Bold} numbers are the best in each Avg. column. We include standard deviations of three independent runs for \ourmethod{} Snake + Rotation.} 
\label{tab:math_generalization}
\vspace{-10pt}
\end{table}
\begin{table}[!t]
    \centering
    \tablestyle{2pt}{1.2}
    \resizebox{\linewidth}{!}{
    \begin{tabular}{l|>{\columncolor{gray!8}}c|
                    >{\columncolor{gray!8}}ccc|
                    >{\columncolor{gray!8}}ccc}
        & \multirow{2}{*}{\textbf{\begin{tabular}{@{}c@{}}\\Avg.\end{tabular}}}
        & \multicolumn{3}{c|}{\textbf{CLEVR$^{+}$}}
        & \multicolumn{3}{c}{\textbf{Multi‐Discipline}}\\
        \multirow{-2}{*}{\textbf{Model}}
        & \textbf{Avg.}
        & \textbf{Avg.} & \textbf{CLEVR‐M} & \textbf{S‐CLEVR}
        & \textbf{Avg.} & \textbf{MMMU$_{\text{val}}$} & \textbf{MMMU‐Pro$_{\text{overall}}$}\\
        \midrule
        \multicolumn{8}{@{}c}{Proprietary Model}\\
        \midrule
        GPT‐4o~\citep{hurst2024gpt}
            & 55.9 & 51.2 & 68.1 & 34.3 & 60.5 & 69.1 & 51.9\\
        Gemini‐2.0‐Flash~\citep{team2023gemini}
            & -- & 46.3 & 64.9 & 27.6 & -- & 71.9 & --\\
        \midrule\midrule
        \multicolumn{8}{@{}c}{Multimodal Reasoning Model Post‐Trained on Qwen2.5‐VL‐7B~\citep{Qwen2.5-VL}}\\
        \midrule
        \textit{Base Model: Qwen2.5‐VL‐7B}
            & 50.3 & 54.9 & 74.6 & 35.2 & 45.7 & 54.3 & 37.0\\
        \midrule
        R1‐Onevision‐7B~\citep{yang2025r1}
            & 53.7 & \textcolor{gray}{65.1} & \textcolor{gray}{75.5} & \textcolor{gray}{54.7}
            & \textcolor{gray}{42.3} & \textcolor{gray}{51.9} & \textcolor{gray}{32.6}\\
        R1‐VL‐7B~\citep{chen2025r1v}
            & 53.9 & \textcolor{gray}{68.0} & \textcolor{gray}{87.4} & \textcolor{gray}{48.6}
            & \textcolor{gray}{39.7} & \textcolor{gray}{50.0} & \textcolor{gray}{29.4}\\
        MM‐Eureka‐Qwen‐7B~\citep{meng2025mm}
            & 62.8 & 79.3 & 98.4 & 60.1
            & 46.4 & 55.8 & 36.9\\
        Reason‐RFT‐Zero‐7B~\citep{tan2025reason}
            & 58.6 & \textcolor{gray}{76.2} & \textcolor{gray}{99.4} & \textcolor{gray}{53.0}
            & 40.9 & 51.2 & 30.6\\
        VLAA‐Thinker‐7B~\citep{chen2025sftrlearlyinvestigation}
            & 61.7 & \textcolor{gray}{\textbf{83.4}} & \textcolor{gray}{94.7} & \textcolor{gray}{72.1}
            & \textcolor{gray}{40.1} & \textcolor{gray}{48.2} & \textcolor{gray}{31.9}\\
        OpenVLThinker‐7B~\citep{deng2025openvlthinker}
            & 60.4 & \textcolor{gray}{82.4} & \textcolor{gray}{93.8} & \textcolor{gray}{71.0}
            & \textcolor{gray}{38.5} & \textcolor{gray}{54.8} & \textcolor{gray}{22.1}\\
        \hline
        \ourmethod{} Snake
            & 64.4 & 82.6 & 92.6 & 72.6
            & 46.2 & 55.8 & 36.6\\
        \ourmethod{} Rotation
            & 63.3 & 80.7 & 93.0 & 68.3
            & 45.9 & 54.1 & 37.7\\
        \ourmethod{} Snake + Rotation
            & \textbf{64.7} & 81.7 & 91.9 & 71.4
            & \textbf{47.7} & 58.0 & 37.4\\
    \end{tabular}}
    \vspace{-5pt}
    \caption{\textbf{Results on multimodal spatial and multi‐discipline reasoning benchmarks.}
    CLEVR‐M denotes CLEVR‐Math~\citep{clevr-math}, and S‐CLEVR stands for Super‐CLEVR~\citep{super-clevr}.
    Results post-trained on the same subject as the evaluation are \textcolor{gray}{de-emphasized}, while \ourmethod{} is exclusively post‐trained using games.
    \textbf{Bold} numbers are the best in each Avg. column.} 
    \label{tab:reason_generalization}
    \vspace{-10pt}
\end{table}
\begin{table}[h]
    \centering
    \tablestyle{2pt}{1.2}
    \resizebox{\linewidth}{!}{
    \begin{tabular}{l|>{\columncolor{gray!8}}c|cccccccc}
        \textbf{Model}\rule{0pt}{2.6ex} & \textbf{Avg.}\rule{0pt}{2.6ex}
        & \rotatebox{30}{\textbf{Obj. Count}} & \rotatebox{30}{\textbf{Abs. Dist.}} & \rotatebox{30}{\textbf{Obj. Size}} & \rotatebox{30}{\textbf{Room Size}}
        & \rotatebox{30}{\textbf{Rel. Dist.}} & \rotatebox{30}{\textbf{Rel. Dir.}} & \rotatebox{30}{\textbf{Route Plan}} & \rotatebox{30}{\textbf{Appr. Order}}\\
        \midrule
        \multicolumn{10}{@{}c}{Proprietary Model}\\
        \midrule
        GPT-4o~\citep{hurst2024gpt} & 34.0 & 46.2 & 5.3 & 43.8 & 38.2 & 37.0 & 41.3 & 31.5 & 28.5 \\
        Gemini-1.5-Flash~\citep{gemini_pro} & 42.1 & 49.8 & 30.8 & 53.5 & 54.4 & 37.7 & 41.0 & 31.5 & 37.8 \\
        Gemini-1.5-Pro~\citep{gemini_pro} & 45.4 & 56.2 & 30.9 & 64.1 & 43.6 & 51.3 & 46.3 & 36.0 & 34.6 \\
        \midrule \midrule
        \multicolumn{10}{@{}c}{Open-source Models}\\
        \midrule
        InternVL2-8B~\citep{chen2024internvl} & 34.6 & 23.1 & 28.7 & 48.2 & 39.8 & 36.7 & 30.7 & 29.9 & 39.6 \\
        InternVL2-40B~\citep{chen2024internvl} & 36.0 & 34.9 & 26.9 & 46.5 & 31.8 & 42.1 & 32.2 & 34.0 & 39.6 \\
        LongVILA-8B~\citep{chenlongvila} & 21.6 & 29.1 & 9.1 & 16.7 & 0.0 & 29.6 & 30.7 & 32.5 & 25.5 \\
        VILA-1.5-40B~\citep{vila} & 31.2 & 22.4 & 24.8 & 48.7 & 22.7 & 40.5 & 25.7 & 31.5 & 32.9 \\
        LongVA-7B~\citep{zhang2024long} & 29.2 & 38.0 & 16.6 & 38.9 & 22.2 & 33.1 & 43.3 & 25.4 & 15.7 \\
        \midrule
        \multicolumn{10}{@{}c}{Multimodal Reasoning Model Post-Trained on Qwen2.5-VL-7B~\citep{Qwen2.5-VL}}\\
        \midrule
        \textit{Base Model (Qwen2.5-VL-7B)} & 36.7 & 41.9 & 21.4 & 50.4 & 36.8 & 38.5 & 40.9 & 29.9 & 34.1 \\
        Video-R1~\citep{feng2025video} & 35.1 & -- & -- & -- & -- & -- & -- & -- & -- \\
        Visual Jigsaw~\citep{wu2025visual} & 38.5 & -- & -- & -- & -- & -- & -- & -- & -- \\
        ViGaL & \textbf{38.7} & 42.9 & 22.6 & 46.7 & 45.8 & 38.3 & 41.7 & 35.1 & 36.1 \\
        \hline
    \end{tabular}}
    \vspace{-5pt}
    \caption{\textbf{Results on VSI-Bench 3D spatial reasoning.} We compare proprietary models, open-source multimodal models, and multimodal reasoning models post-trained on Qwen2.5-VL-7B.}
    \label{tab:spatial_generalization}
    \vspace{-10pt}
\end{table}

\section{Visual Reasoning Generalization}
\label{sec:exp}

\paragraph{Evaluation collection.}
\label{sec:eval_data}
Following prior studies~\citep{tong2024cambrian, li2024survey}, we systematically divide existing benchmarks into two broad categories: (i) \emph{reasoning-oriented benchmarks} requiring multi-step or mathematical reasoning, and (ii) \emph{general-purpose perception benchmarks} assessing visual understanding and perception abilities.

For reasoning-oriented evaluation, we test on four key areas: \textit{Math} (MathVista~\citep{lu2024mathvistaevaluatingmathematicalreasoning}, MathVerse~\citep{zhang2024mathversedoesmultimodalllm}, MathVision~\citep{wang2024measuring}), \textit{Geometry} (GeoMath~\citep{geo170k, shi2024mathllavabootstrappingmathematicalreasoning}, Geometry3K~\citep{lu2021inter}), \textit{CLEVR+} (CLEVR-Math~\citep{clevr-math}, Super-CLEVR~\citep{super-clevr}), and \textit{Multi-Discipline} (MMMU~\citep{yue2024mmmu}, MMMU-Pro~\citep{yue2024mmmupro}). For general perception, we evaluate across three categories: General (MuirBench~\citep{wang2024muirbench}, CRPE~\citep{kazemzadeh2014referitgame}), Vision-Centric (MMVP~\citep{tong2024eyes}, RealWorldQA~\citep{grok15}, MMStar~\citep{chen2024we}, MME~\citep{fu2023mme}, BLINK~\citep{fu2024blink}), and OCR \& Chart (AI2D~\citep{kembhavi2016diagram}, SEED-Bench-2-Plus~\citep{li2024seed2plus}, DocVQA~\citep{docvqa}, OCRBench~\citep{liu2024ocrbenchhiddenmysteryocr}). More detailed descriptions of each benchmark are provided in Appendix~\ref{sec:detailed_eval_benchmarks}.

\subsection{Main Results}
\label{sec:main_results}


\paragraph{Zero-shot generalization from gameplay to multimodal reasoning.} Our approach consistently shows remarkable generalization capabilities on mathematical and other reasoning tasks, despite having no direct exposure to in-domain training data during RL post-training. As shown in Tab.~\ref{tab:math_generalization}, our method notably outperforms models specifically RL-trained on mathematical tasks. For instance, \ourmethod{} Snake + Rotation achieves 0.5\% higher accuracy than MM-Eureka-Qwen-7B~\citep{meng2025mm} 28.7\% on Geometry, even though MM-Eureka-Qwen-7B was explicitly trained on high-quality mathematical and geometry datasets.

This strong generalization extends beyond mathematics. Tab.~\ref{tab:reason_generalization} shows that \ourmethod{} Snake + Rotation outperforms R1-OneVision-7B~\citep{yang2025r1} by 5.4\% on average across MMMU series benchmarks, which test multi-disciplinary reasoning. This is particularly notable since R1-OneVision-7B was trained on a carefully curated comprehensive dataset spanning multiple subjects.

Furthermore, Tab.~\ref{tab:spatial_generalization} demonstrates that post-training on two games using synthesized 2D images successfully generalizes to 3D spatial reasoning tasks. On VSI-Bench~\citep{yang2025thinking}, \ourmethod{} improves by 2.0\% after game-based posttraining, slightly surpassing Visual Jigsaw~\citep{wu2025visual}. This is notable because Visual Jigsaw is posttrained on 3D video data, whereas our approach uses only 2D game data.
\begin{figure}[t]
    \centering
    \begin{subfigure}[t]{0.48\linewidth}
        \centering
        \includegraphics[width=\linewidth]{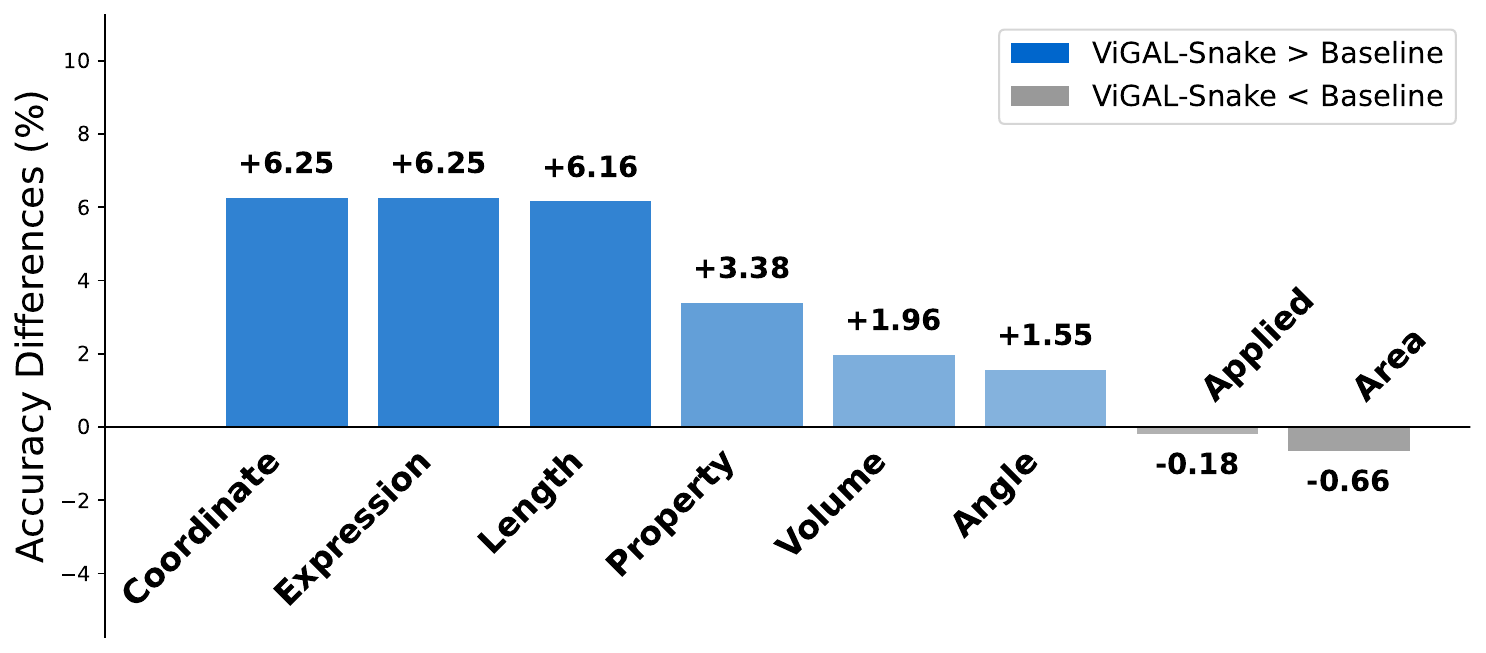}
        \caption{Snake vs. Baseline}
    \end{subfigure}
    \hfill
    \begin{subfigure}[t]{0.48\linewidth}
        \centering
        \includegraphics[width=\linewidth]{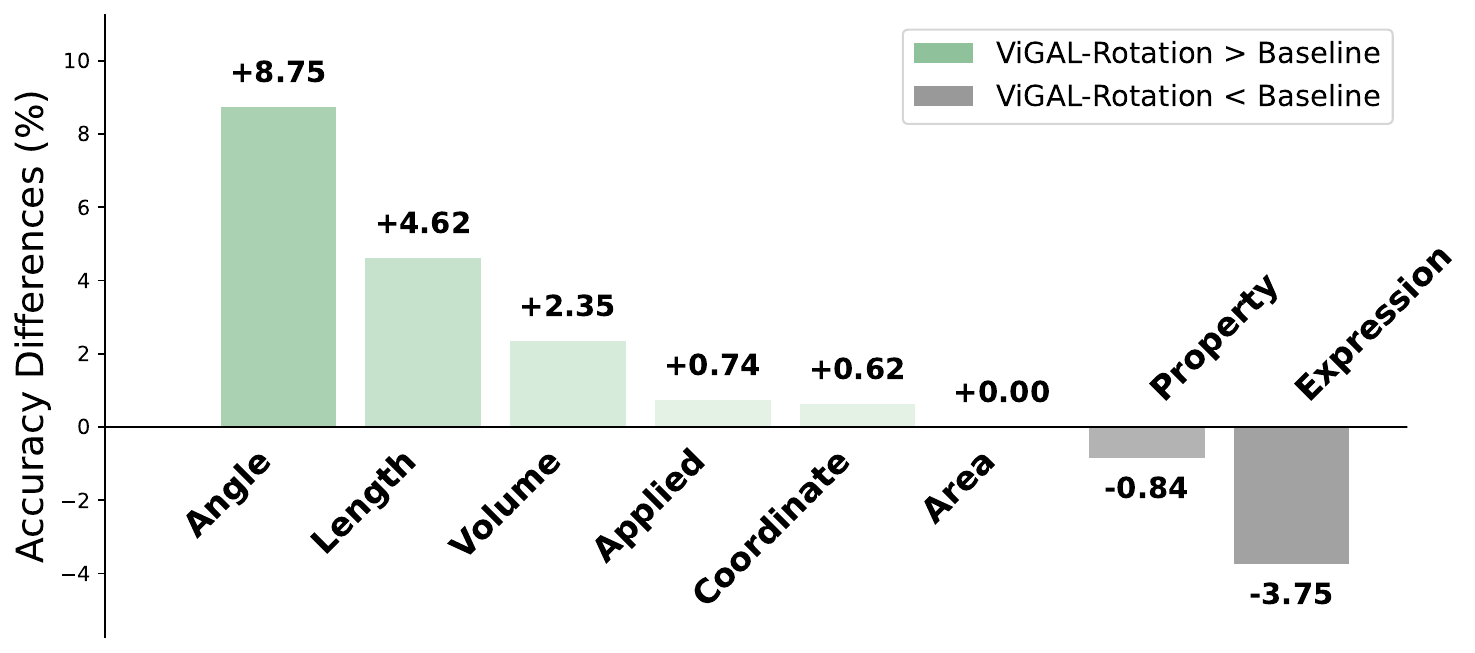}
        \caption{Rotation vs. Baseline}
    \end{subfigure}
    \caption{\textbf{Per-category gains on MathVerse are \textit{not} uniform.} The eight math categories follow MathVerse~\citep{zhang2024mathversedoesmultimodalllm}. (a) Snake yields the largest gains on \textit{Coordinates} and \textit{Expressions}, consistent with its 2D grid structure. (b) Rotation boosts \textit{Angle} and \textit{Length} questions but reduces \textit{Expression} accuracy, suggesting its training primarily incentivizes orientation recognition.}
    \label{fig:acc_diff}
    \vspace{-10pt}
\end{figure}

\begin{figure}[t]
    \centering
    \includegraphics[width=0.99\linewidth]{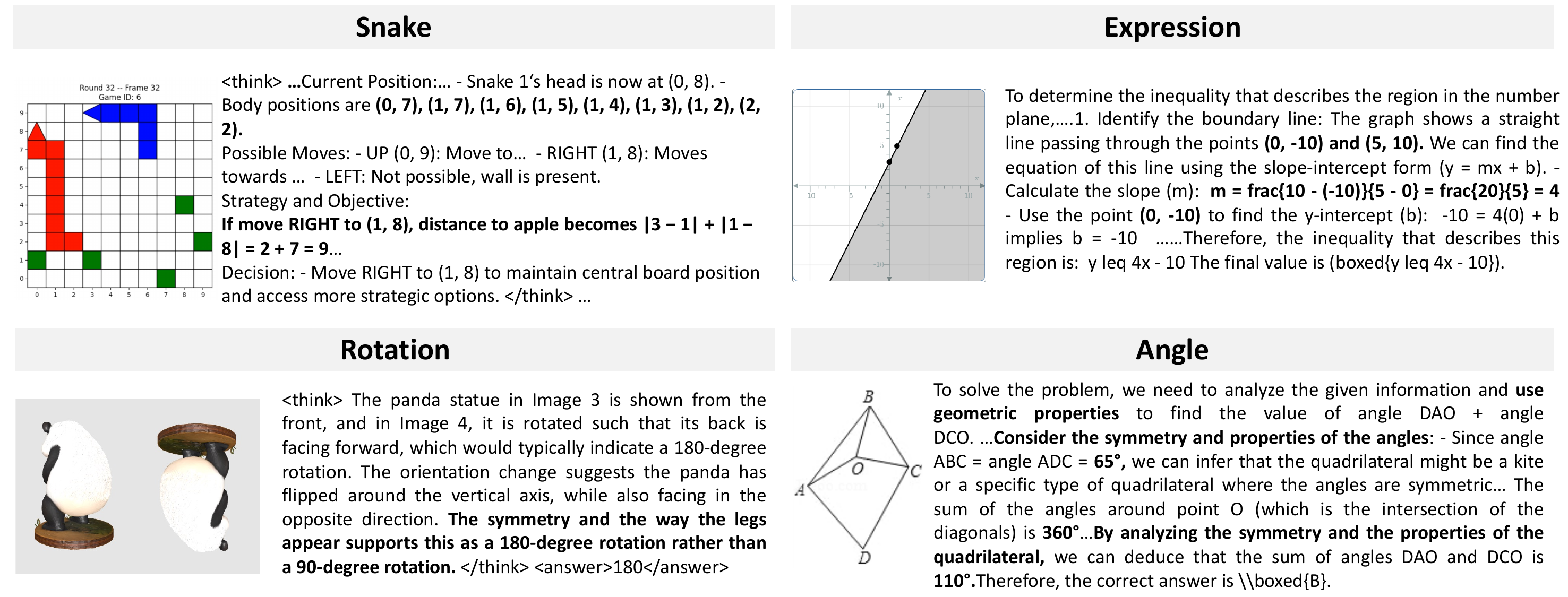}
    \caption{\textbf{Reasoning trace of different games and math questions.} Top: Algebraic functions and coordinate-level interpretations that emerge from playing the Snake game help solving \textit{Expression} questions. Bottom: Spatial reasoning skills incentivized by playing the Rotation game appear when solving \textit{Angle}-related problems.
    }
    \vspace*{-0.5cm}
    \label{fig:game_help_math}
\end{figure}

These empirical results suggest that gameplay-based post-training develops fundamental reasoning capabilities that transfer more effectively than direct RL training on diverse task-specific datasets. Moreover, the gameplay environment appears to encourage general problem-solving strategies that consistently generalize well to out-of-domain tasks.

\paragraph{Blending multiple games enhances generalization.}
As shown in Tab.~\ref{tab:math_generalization}, post-training on Snake achieves best performance on the CLEVR+ benchmark, while training on Rotation yields stronger results on geometry reasoning. Furthermore, training on both Snake and Rotation enables learning complementary skills, improving the overall benchmark average to 63.1\%. These findings indicate that combining game environments drives meaningful performance gains, demonstrating Visual Gaming Learning's potential as a promising paradigm for enhancing generalizable reasoning without large-scale domain-specific data. Expanding the types of games during training consistently scales performance across different visual reasoning tasks.

\paragraph{Different games benefit distinct math subfields.}
\label{sec:acc_diff}
To study which types of problems in the math benchmarks benefit from game play, we analyze accuracy differences across MathVerse~\citep{zhang2024mathversedoesmultimodalllm} subcategories between \ourmethod models trained with Snake or Rotation, as shown in Fig.~\ref{fig:acc_diff}. We find that training on the Snake game significantly improves performance on the subcategories like Expressions and Coordinates, while training on Rotation notably enhances performance on questions about angles and lengths. To understand why different games help with different types of math, we compare the reasoning processes required for playing games versus solving math problems. As shown in Fig.~\ref{fig:game_help_math}, solving Expressions questions involves algebraic functions and coordinate-level interpretations of graphical representations, which closely align with the spatial reasoning process in Snake. Similarly, solving angle-related questions is consistent with requirement of playing Rotation game to reason about rotational angles of 3D objects. These results suggest that playing different games develops fundamental skills like spatial modeling and algebraic calculation that transfer to visual math questions. The experiment on quantitatively analyzing the correlation between math and game is in Sec.~\ref{sec:game_math_rel} in the Appendix. Furthermore, joint training on both games leads to improvements across \textit{all} reasoning categories (see Appendix Sec.~\ref{sec:acc_diff_appendix}). We also include qualitative analyses on improvements in math reasoning after RL in Appendix Sec.~\ref{sec:case_study}.

\begin{table}[t]
  \centering
  \tablestyle{4.5pt}{1.2}
  \small
  \resizebox{0.8\linewidth}{!}{
  \begin{tabularx}{0.85\linewidth}{lXXXX}
      \textbf{Model} & \textbf{Avg.} & \textbf{MathVista} & \textbf{MathVerse} & \textbf{MathVision} \\ \midrule
      \textcolor[gray]{0.5}{Base Model: Qwen2.5-VL-7B} & \textcolor{gray}{47.7} & \textcolor{gray}{68.0} & \textcolor{gray}{49.0} & \textcolor{gray}{26.0} \\
      MM-Eureka-Qwen-7B & 50.1 & 73.0 & 50.3 & 26.9 \\
      \ourmethod{} Snake + Rotation & 50.6 & 71.9 & 52.4 & 27.5 \\
      \mycolor{\ourmethod{} Snake + Rotation + Math Data} & \mycolor{51.8} & \mycolor{72.3} & \mycolor{54.5} & \mycolor{27.7} \\
\end{tabularx}}
\vspace{-5pt}
\caption{\textbf{Gameplay complements math data.} Adding math data MMK12 on top of \ourmethod{} yields further gains in math performance. With access to the same amount of math data, \ourmethod{} outperforms MM-Eureka~\citep{meng2025mm} on average of the three math benchmarks.}
\label{tab:withmath}
\vspace{-10pt}
\end{table}
\paragraph{Gameplay complements math data.}
We explore the complementary benefits of adding math data to the gameplay training pipeline, for which we implement a two-stage training process. Stage 1 equals to \ourmethod{} setup, training the model on Snake and Rotation games. In stage 2, we further finetune the stage 1 model on MMK12~\citep{meng2025mm}, a multimodal mathematical reasoning dataset containing approximately 12k examples. Stage 2 training uses the identical data and settings as MM-Eureka-Qwen-7B~\citep{meng2025mm}.  As shown in Tab.~\ref{tab:withmath}, the integration of mathematical data in stage 2 yields a continuous improvement of 1.2\% on average across three mathematical benchmarks. This demonstrates the complementary relationship between our visual game learning approach and mathematical data post-training. Moreover, \ourmethod{} with math data significantly outperforms MM-Eureka-Qwen-7B by 1.7\% on mathematical benchmarks on average, using the same math data. These results suggest that visual game learning can serve as an effective surrogate task together with domain-specific data to improve performance on target tasks.

\begin{table}[t]
    \centering
    \tablestyle{1pt}{1.0}
    \resizebox{\linewidth}{!}{
    \begin{tabular}{l|>{\columncolor{gray!8}}c|
                        >{\columncolor{gray!8}}ccc|
                        >{\columncolor{gray!8}}ccccc|c|
                        >{\columncolor{gray!8}}ccccc}
        & \multirow{2}{*}{\textbf{\begin{tabular}{@{}c@{}}\\Avg.\end{tabular}}}
        & \multicolumn{3}{c|}{\textbf{General}}
        & \multicolumn{6}{c|}{\textbf{Vision-Centric}}
        & \multicolumn{5}{c}{\textbf{OCR \& Chart}}\\
        \multirow{-2}{*}{\textbf{Model}}
        & \textbf{Avg.} 
        & \textbf{Avg.}
        & \textbf{\begin{tabular}{c}Muir-\\Bench\end{tabular}}
        & \textbf{CRPE$_{\text{rel.}}$}
        & \textbf{Avg.}
        & \textbf{MMVP}
        & \textbf{\begin{tabular}{c}Real-\\WorldQA\end{tabular}}
        & \textbf{MMStar}
        & \textbf{BLINK$_{\text{val}}$}
        & \textbf{MME$_{\text{p}}$}
        & \textbf{Avg.}
        & \textbf{\begin{tabular}{c}AI2D\\$_{\text{w.\,M.}}$\end{tabular}}
        & \textbf{\begin{tabular}{c}SEED-\\Bench-2+\end{tabular}}
        & \textbf{{\begin{tabular}{c}DocVQA\\{val}\end{tabular}}}
        & \textbf{\begin{tabular}{c}OCR-\\Bench\end{tabular}}\\
        \midrule
        \multicolumn{16}{@{}c}{Proprietary Model}\\
        \midrule
        GPT-4o~\citep{hurst2024gpt}               & 74.8 & 72.3 & 68.0 & 76.6 & 69.4 & --   & 75.4 & 64.7 & 68.0 & 1614 & 82.6 & 84.6 & 72.0 & 91.1 & 736\\
        \midrule
        \multicolumn{16}{@{}c}{General Multimodal Language Model}\\
        \midrule
        Qwen2.5-VL-7B~\citep{Qwen2.5-VL}
                                                & 72.4 & 68.0 & 59.6 & 76.4 & 65.8 & 74.3 & 68.5 & 63.9 & 56.4 & 1698 & 83.3 & 83.9 & 70.4 & 95.7 & 864\\
        \midrule
        \multicolumn{16}{@{}c}{Multimodal Reasoning Model Post-Trained on Qwen2.5-VL-7B}\\
        \midrule
        R1-Onevision-7B~\citep{yang2025r1}       & --   & 66.8 & 46.3 & 87.3 & 56.5 & 61.3 & 58.0 & 57.8 & 48.7 & 1504 & --   & --   & --   & --   & --\\
        R1-VL-7B~\citep{chen2025r1v}             & 67.4 & 63.3 & 54.1 & 72.4 & 59.6 & 70.3 & 61.4 & 55.6 & 51.0 & 1657 & 79.2 & 81.7 & 66.4 & 89.4 & 81.0\\
        MM-Eureka-Qwen-7B~\citep{meng2025mm}     & 71.8 & \textbf{68.9} & 61.1 & 76.7 & 65.1 & 74.3 & 66.1 & 65.9 & 54.0 & 1626 & 81.5 & 84.3 & 68.2 & 92.0 & 87.0\\
        Reason-RFT-Zero-7B~\citep{tan2025reason} & 68.4 & 66.9 & 58.5 & 75.2 & 58.5 & 58.0 & 65.3 & 59.1 & 51.6 & 1653 & 79.8 & 83.3 & 68.0 & 88.1 & 82.0\\
        VLAA-Thinker-7B~\citep{chen2025sftrlearlyinvestigation}
                                                & 69.7 & 65.9 & 57.1 & 74.6 & 62.6 & 71.6 & 65.4 & 60.4 & 53.0 & 1593 & 80.6 & 83.4 & 67.4 & 90.9 & 84.5\\
        OpenVLThinker-7B~\citep{deng2025openvlthinker}
                                                & --   & 64.3 & 52.8 & 75.8 & 50.4 & 32.3 & 60.2 & 59.1 & 49.9 & 1513 & --   & --   & --   & --   & --\\
        \hline
        \ourmethod{} Snake + Rotation           & \textbf{72.2} & 68.6 & 60.5 & 76.7 & \textbf{65.7} & 74.6 & 67.3 & 65.4 & 55.6 & 1685 & \textbf{82.2} & 84.8 & 69.1 & 92.7 & 86.6\\
    \end{tabular}}
    \caption{\textbf{Main results on multimodal language benchmarks targeting more general and comprehensive visual ability.} We compare with models post-trained on Qwen2.5-VL-7B~\citep{Qwen2.5-VL}.  Best category averages are highlighted in \textbf{bold}. Note that MME$_{\text{p}}$ is excluded from vision-centric category average accuracy due to scale differences.}
    \label{tab:vision_generalization}
    \vspace{-5pt}
\end{table}
\paragraph{Preserving general visual capabilities while reasoning enhancement.}

To examine whether generalization on reasoning tasks leads to degradation in general visual capabilities, we evaluate \ourmethod{} Snake + Rotation on a broader set of MLLM benchmarks. As shown in Tab.~\ref{tab:vision_generalization}, compared to Qwen2.5-VL-7B prior to RL tuning, our model maintains comparable general visual performance while achieving stronger math reasoning results. In contrast, other models that improve math performance through RL post-training often exhibit substantial drops in general visual capabilities. These results demonstrate that our gameplay-based approach enables math generalization without compromising other visual abilities.

\vspace{-5pt}

\begin{table}[t]
\centering

\begin{subtable}[t]{0.32\textwidth}
\centering
\tablestyle{4.5pt}{1.2}
\caption{\centering Text prompt design.}

\label{tab:ablation_prompt}
\resizebox{\linewidth}{!}{
\begin{tabular}{y{100}cccc}
prompt & \textbf{Avg.} & \textbf{Math} & \textbf{CLEVR+} & \textbf{Geo.} \\ 
\midrule 
\textcolor[gray]{0.5}{base model} & \textcolor{gray}{49.1} & \textcolor{gray}{47.7} & \textcolor{gray}{54.9} & \textcolor{gray}{44.8} \\ 
w/o reasoning instr. & 59.5 & 48.0 & 80.4 & 50.1 \\ 
\mycolor{w/ reasoning instr.} & \mycolor{\textbf{62.3}} & \mycolor{49.4} & \mycolor{82.6} & \mycolor{55.0} \\  
\end{tabular}
}
\end{subtable}
\def\graycolor{\cellcolor[HTML]{eaeaea}}
\begin{subtable}[t]{0.32\textwidth}
  \centering
  \tablestyle{4.5pt}{1.2}
  \caption{\centering Reward design.}
  \label{tab:ablation_reward}
  \resizebox{\linewidth}{!}{
    \begin{tabular}{y{100}cccc}
      reward                       & \textbf{Avg.}          & \textbf{Math}       & \textbf{CLEVR+}     & \textbf{Geo.}       \\ 
      \midrule
      {\textcolor{gray}{base model}}        & {\textcolor{gray}{49.1}} & {\textcolor{gray}{47.7}} & {\textcolor{gray}{54.9}} & {\textcolor{gray}{44.8}} \\ 
      best moves                   & 59.6                  & 48.2               & 80.4               & 50.2               \\ 
      {\mycolor best \& worst moves}  & {\mycolor \textbf{62.3}}         & {\mycolor 49.4}        & {\mycolor 82.6}        & {\mycolor 55.0}        \\ 
      {\graycolor \hspace{1em} w/ random label} & {\graycolor 49.4}      & {\graycolor 47.5}    & {\graycolor 55.4}    & {\graycolor 47.5}    \\ 
    \end{tabular}
  }
\end{subtable}
\begin{subtable}[t]{0.32\textwidth}
\centering
\tablestyle{4.5pt}{1.2}
\caption{\centering Difficulty control.}
\label{tab:ablation_difficulty}
\resizebox{\linewidth}{!}{
\begin{tabular}{y{100}cccc}
difficulty control & \textbf{Avg.} & \textbf{Math} & \textbf{CLEVR+} & \textbf{Geo.} \\ 
\midrule 
\textcolor[gray]{0.5}{base model} & \textcolor{gray}{49.1} & \textcolor{gray}{47.7} & \textcolor{gray}{54.9} & \textcolor{gray}{44.8} \\ 
w/o  difficulty control & 60.6 & 48.8 & 81.4 & 51.8 \\ 
\mycolor{w/ difficulty control} & \mycolor{\textbf{62.3}} & \mycolor{49.4} & \mycolor{82.6} & \mycolor{55.0} \\ 
\end{tabular}
}
\end{subtable}

\begin{subtable}[t]{0.32\textwidth}
\centering
\tablestyle{4.5pt}{1.2}
\caption{\centering Data scalability.}
\label{tab:ablation_scaling}
\resizebox{\linewidth}{!}{
\begin{tabular}{y{100}cccc}
training samples & \textbf{Avg.} & \textbf{Math} & \textbf{CLEVR+} & \textbf{Geo.} \\ 
\midrule 
\textcolor[gray]{0.5}{base model} & \textcolor{gray}{49.1} & \textcolor{gray}{47.7} & \textcolor{gray}{54.9} & \textcolor{gray}{44.8} \\ 
16K & 60.1 & 48.9 & 81.2 & 50.3 \\ 
\mycolor{36K} & \mycolor{\textbf{62.3}} & \mycolor{49.4} & \mycolor{82.6} & \mycolor{55.0} \\ 
\end{tabular}
}
\end{subtable}
\begin{subtable}[t]{0.32\textwidth}
\centering
\tablestyle{4.5pt}{1.2}
\caption{\centering Input modality.}
\label{tab:ablation_vision}
\resizebox{\linewidth}{!}{
\begin{tabular}{y{100}cccc}
input modality & \textbf{Avg.} & \textbf{Math} & \textbf{CLEVR+} & \textbf{Geo.} \\ 
\midrule 
\textcolor[gray]{0.5}{base model} & \textcolor{gray}{49.1} & \textcolor{gray}{47.7} & \textcolor{gray}{54.9} & \textcolor{gray}{44.8} \\ 
text & 59.6 & 48.5 & 80.1 & 50.3 \\ 
\mycolor{vision \& text} & \mycolor{\textbf{62.3}} & \mycolor{49.4} & \mycolor{82.6} & \mycolor{55.0} \\ 
\end{tabular}
}
\end{subtable}
\begin{subtable}[t]{0.32\textwidth}
\centering
\tablestyle{4.5pt}{1.2}
\caption{\centering SFT vs. RL.}
\label{tab:ablation_sftvsrl}
\resizebox{\linewidth}{!}{
\begin{tabular}{y{100}cccc}
post-training & \textbf{Avg.} & \textbf{Math} & \textbf{CLEVR+} & \textbf{Geo.} \\ 
\midrule 
\textcolor[gray]{0.5}{base model} & \textcolor{gray}{49.1} & \textcolor{gray}{47.7} & \textcolor{gray}{54.9} & \textcolor{gray}{44.8} \\ 
SFT & 47.2 & 38.0 & 71.5 & 32.1 \\
\mycolor{RL} & \mycolor{\textbf{62.3}} & \mycolor{49.4} & \mycolor{82.6} & \mycolor{55.0} \\ 

\end{tabular}
}
\end{subtable}

\vspace{-5pt}
\caption{\textbf{Ablation study.} We ablate different aspects of \ourmethod with Snake and evaluate on downstream benchmarks. The similar evaluation with Rotation is in Sec.~\ref{sec:ablation_rot}. Each benchmark consists of several subtasks (Tab.~\ref{tab:math_generalization} and Tab.~\ref{tab:reason_generalization}), and we report their averages. The base model is Qwen2.5-VL-7B, whose results are in \textcolor[gray]{0.5}{gray}. The default settings in Tab.~\ref{tab:math_generalization} and Tab.~\ref{tab:reason_generalization} are highlighted in \colorbox[HTML]{CFEFFF}{blue}.}
 \vspace*{-0.5cm}
\label{tab:ablation}
\end{table}

\subsection{Ablation Study}
\label{sec:ablation}
We ablate key design choices in the Snake environment, evaluate each variant on downstream benchmarks, and report the results in Tab.~\ref{tab:ablation}. The corresponding ablation for the Rotation environment is provided in Appendix Sec.~\ref{sec:ablation_rot}.

\paragraph{Reasoning instructions in the text prompt help.}
We use reasoning instructions, such as ``\texttt{\small finding the nearest apple by calculating Manhattan distances}'', in the text prompts to guide the model thinking chains. The complete text prompts are in Appendix Sec.~\ref{sec:prompt}. In Tab.~\ref{tab:ablation_prompt}, we demonstrate that reasoning instructions brings significant improvement of 1.9\%, from 59.5\% to 61.4\%, for Snake in average accuracy over the three out-of-domain benchmarks.

\paragraph{Reward design of pre-text game matters for downstream tasks.}
We show that reward design of RL for games plays a crucial role for the downstream tasks. As shown in Tab.~\ref{tab:ablation_reward}, we first ask the model to predict only the best next move, defined as the action that moves toward the closest apple while avoiding death. In our improved reward design, we task the model with simultaneously predicting both the best and worst next moves, where the worst move leads directly to losing the game. More importantly, it leads to improvements across all downstream tasks, bringing an average increase of 1.8\%. These results suggest that proper reward design in pre-text game can improve not only gameplay capabilities but also generalization to downstream tasks.

Furthermore, inspired by several prior works that improve model performance without labeled rewards~\citep{zhao2025learning} or with random labels~\citep{shao2025spurious}, we also provide a random reward ablation, where we still ask the model to predict both best and worst moves but use random moves as the labels. We report the results in the last row in Tab.~\ref{tab:ablation_reward}. In our gameplay setting, RL with random labels reports 49.4\% on averagne and does no provide significant gains over the base model, different from the conclusions in prior works~\citep{shao2025spurious}. Potential explanations lie in the difference in data domains and base models, where other works applied random labels to text-only mathematical data while our work applies random labels to visual game data.

\paragraph{Controlling game difficulty for better reasoning.}
Gameplay for RL post-training offers unique opportunities to easily control task difficulty. We present an ablation study on difficulty control importance. We define difficulty based on snake length, where longer snakes represent higher difficulty. For controlled difficulty, we collect training data using states where snake length falls within a moderate range of 1-5. Details are in Sec.~\ref{sec:data_appendix}. As shown in Tab.~\ref{tab:ablation_difficulty}, difficulty control achieves 61.4\% overall accuracy compared to 60.6\% without control. This suggests our game engine can easily generate appropriately difficult data, helping prevent model sub-optimization during RL training.

\paragraph{RL on games shows data scalability.} 
Thanks to using game engine, we can generate data at any scale with high flexibility. To show data scalability on RL of visual games, we conduct experiments using 16k and 32k snake game samples, respectively. As in Tab.~\ref{tab:ablation_scaling}, scaling data from 16k to 32k brings a performance improvement of 1.3\% on average across all domains. This suggests the potential of the proposed ViGaL paradigm to improve downstream performance by easily scaling training data, which contrasts with the data scaling challenges of domain-specific human annotated data, requiring extensive manual effort.

\paragraph{Both text and vision contribute to better visual reasoning.}
To isolate the contributions of text and vision modalities, we conduct an ablation study with a text-only setting. In this setup, we represent game states—including snake positions, apple locations, and boundary constraints—using only textual descriptions during RL training. The model trained with text-only inputs on the Snake game demonstrates substantial improvements across all multimodal benchmarks, with average performance increasing from 49.1\% to 59.6\%. Incorporating visual inputs yields an additional 1.8\% performance gain. These results demonstrate that multimodal RL enhances visual reasoning capabilities, with complementary contributions from both text and vision modalities. 

\vspace{-2pt}
\paragraph{RL generalizes better than SFT from games to math.}
To evaluate the out-of-domain generalization of \ourmethod, we compare it with supervised fine-tuning (SFT) using identical visual game data. Tab.~\ref{tab:ablation_sftvsrl} shows that SFT with Snake game data degrades the base model's performance on both mathematical reasoning and geometry tasks by a notable 9.7\% and 12.7\%, respectively. While SFT produces modest improvements on CLEVR+, these gains are substantially smaller than those achieved by RL. Overall, RL improves performance by 12.3\%, whereas SFT decreases performance by 1.9\%. This stark contrast demonstrates that RL better preserves and extends the model's reasoning capabilities to new domains.

\vspace{5pt}
\section{Related Work}
\label{sec:related}
\paragraph{Reinforcement Learning in MLLMs.} Reinforcement Learning (RL) increasingly enhances reasoning in Large Language Models (LLMs) beyond Supervised Fine-Tuning (SFT). Text-only models like DeepSeek-R1~\citep{deepseekai2025} demonstrate RL's efficacy, especially with rule-based rewards, for complex reasoning. This paradigm is now being extended to Multimodal LLMs (MLLMs). Recent MLLM research explores RL for improved visual reasoning, drawing from LLM successes. Various works~\citep{peng2025lmmr1, huang2025vision, chen2025r1v} investigate multi-stage training, trace supervision, or rule-based RL for specific visual subdomains like geometry and counting. Others focus on different RL algorithms like Process Reward Models (PRMs)~\citep{luo2025ursa,xiang2024atomthink}, often moving beyond SFT-based Chain-of-Thought generation~\citep{dong2024insight,thawakar2025llamav}. Many efforts favor simpler rule-based rewards~\citep{huang2025vision,zhou2025r1} over complex reward models prone to hacking~\citep{eisenstein2023helping}. Unlike approaches training on costly, domain-specific reasoning datasets, our ViGaL paradigm extends rule-based RL to simple, synthetic visual games, demonstrating these serve as scalable, cost-effective pre-text tasks.

\paragraph{Generalization in MLLMs.} Achieving robust generalization to novel tasks, distributions, and domains is central to MLLM development. RL shows promise for better out-of-distribution (OOD) generalization compared to SFT~\citep{chen2025r1v,meng2025mm}, and developing multi-step reasoning like CoT~\citep{wei2022chain} is itself generalization. This is often pursued through training on large, diverse instruction-following datasets~\citep{li2024llavanext,liu2024llavanext,chen2024internvl} or explicitly training general reasoning capabilities~\citep{yang2025r1,huang2025vision}. While these methods advance OOD generalization, they typically operate within the same broad domain of complex visual reasoning as training data.The work most closely related to ours is Game-RL~\citep{tong2025code2logic}, which demonstrates how constructing diverse game-based tasks for RL can enhance the general reasoning abilities of vision-language models (VLMs). Our ViGaL paradigm investigates stronger out-of-domain generalization, showing fundamental skills learned from simple synthetic games transfer zero-shot to enhance performance on entirely different, complex domains like visual mathematics, multi-discipline questions, and spatial reasoning, without domain-specific data exposure.
\vspace{-5pt}
\section{Conclusion}
\vspace{-8pt}
\label{sec:conclusion}
We introduced Visual Game Learning (ViGaL), a novel post-training paradigm where MLLMs learn transferable reasoning by playing simple arcade-style games. Our core finding is that RL on games like Snake and Rotation, \textit{without any in-domain math data}, significantly boosts MLLM performance on mathematical and multi-discipline benchmarks, surpassing specialized models and even large proprietary systems. Ablations confirm the importance of game design, reward structure, and that RL outperforms SFT, while distinct games unlock different skills. We posit that games instill fundamental reasoning skills, suggesting a new avenue for using scalable, controllable synthetic games as powerful pre-text tasks to unlock generalizable reasoning. This work opens doors to exploring a broader range of game-based learning for generalizable AI. 

\bibliography{arxiv} 

\newpage
\appendix
\enlargethispage*{0pt}
\section*{Appendix}
\noindent\textbf{Content}

\noindent A. Data\dotfill 19

\hyperlink{A1}{A.1. Training Data Synthesis \dotfill 19}

\hyperlink{A2}{A.2. Training Prompt in Visual Game Learning \dotfill 20}

\hyperlink{A3}{A.3. Detail of Format Reward \dotfill 22}

\noindent B. Evaluation \dotfill 23

\hyperlink{B1}{B.1. Evaluation Detail of Atari Game\dotfill 23}

\hyperlink{B2}{B.2. Ablation On Rotation Game \dotfill 24}

\hyperlink{B3}{B.3. Synergistic Effects of Multi-Game Training \dotfill 25}

\hyperlink{B4}{B.4.  Reasoning Ability Boundary via Pass$@k$ Evaluation\dotfill 26}

\hyperlink{B5}{B.5. Detail of Evaluation Benchmarks \dotfill 27}

\hyperlink{B6}{B.6. Inference Length Analysis \dotfill 28}

\hyperlink{B7}{B.7. Reasoning Correlation Analysis Between Game and Math \dotfill 28}

\noindent \hyperlink{C}{C. Case Study \dotfill 30}

\newpage

\section{Data}
\hypertarget{A1}{}\subsection{Training Data Synthesis}
\label{sec:data_appendix}

Thanks to using the synthetic game data engine, we can flexibly generate large-scale training data with precisely controlled difficulty levels. 
This completely eliminates the need for extensive data filtering strategies used in previous rule-based RL work training on domain-specific data like math~\citep{meng2025mm,bae2025online}, where difficulty is hard to define and filtering can significantly reduce dataset size. 

For the Snake game, the environment consists of a $10 \times 10$ grid game board with two snakes of 1-grid initial length, where at each time step $t$, each snake receives one action to move, resulting in a new game state $s_{t+1}$. 
We define difficulty based on snake length—longer snakes create more complex game situations and more constrained movement options, closely aligning with how humans perceive difficulty when playing Snake. 
To generate meaningful moves that accomplish the objective of collecting more apples while remaining alive, we implement a policy network based on Proximal Policy Optimization (PPO)~\citep{schulman2017proximal}. 
The observation space is represented as a $10 \times 10$ grid with distinct values indicating empty cells (0), apples (1), the agent's own body (2), and other agents' bodies (3), stacked across 4 time steps to incorporate temporal information, resulting in an input tensor $\mathbf{X} \in \mathbb{R}^{10 \times 10 \times 4}$. 
The policy network architecture consists of two convolutional layers with $3 \times 3$ kernels ($C_1 = 16$ and $C_2 = 32$ output channels), followed by fully connected layers that output action logits for the four possible movements, transformed into a probability distribution $\boldsymbol{\pi}(a|s)$ using softmax. 
To prevent suicidal moves, we incorporate action priors by masking logits for dangerous actions. 
The model employs the standard PPO objective with entropy regularization coefficient $\beta = 0.01$, value function coefficient $\lambda = 0.5$, and clipping parameter $\varepsilon = 0.2$. 
Agents receive rewards of $r = +1$ for collecting apples and penalties of $r = -1$ for dying, enabling them to learn complex behaviors such as obstacle avoidance, apple pursuit, and multi-step trajectory planning. 

For the Rotation game, training data comprises synthetically generated visual puzzles focused on 3D spatial reasoning, utilizing 540 unique 3D object meshes (408 from Hunyuan3D 2.0~\citep{hunyuan3d22025tencent} and 132 from Hunyuan3D 2.5). 
Our custom pipeline produces pairs of images ($I_{\text{init}}, I_{\text{rot}}$) representing objects before and after defined rotations. 
Difficulty in Rotation is determined by the rotation angle between two images, where smaller angle differences present greater perceptual challenges. 
Each pair is generated through a precise sequence: establishing diverse initial viewpoints through compound transformations (base orientation plus additional z-axis rotation from $\{0^\circ, 30^\circ, \dots, 330^\circ\}$ to prevent trivial pattern learning), then applying target rotations of either $90^{\circ}$ or $180^{\circ}$ exclusively around the z-axis. 
All objects are rendered at $512 \times 512$ pixel resolution using a consistent perspective camera under standardized lighting conditions, resulting in approximately 32k unique pairs. 
\enlargethispage*{0pt}

Based on empirical results, we established optimal difficulty parameters for RL training across both games, which we ablate in Tab.~\ref{tab:ablation_difficulty}. 
This controlled progression of difficulty, made possible by our synthetic data generation approach, enables more effective learning trajectories compared to traditional data collection methods.
\hypertarget{A2}{}\subsection{Training Prompt in Visual Game Learning}
\enlargethispage*{0pt}
\begin{figure}[H]
    \centering
\input{tables/prompt_snake}
\end{figure}
\begin{figure}[H]
    \centering
\input{tables/prompt_rotation}
\end{figure}
\label{sec:prompt_appendix}

While the model takes images as input to understand the current state of the game, we design a structural text prompt framework to also provide game guidance. Our game prompts consist of two parts: (1) game settings and (2) reasoning instructions. (1) To help the model understand the game environment, we describe the background, current game state, rules, goals, action space, \textit{etc}. in text besides the input image. (2) In the reasoning instruction part, we provide specific thinking guidance since games can be approached with various thinking chains. To encourage broader thinking, we implement different types of reasoning instructions to guide decision-making process. Specifically, we used GPT-4o~\citep{hurst2024gpt} to synthesize mathematical thinking instructions for Snake, such as ``\texttt{finding the nearest apple by calculating Manhattan distances}'', and spatial thinking instructions for Rotation, for example, ``\texttt{identify major symmetry axes in the original image}''. With reasoning instructions for games, the obtained reasoning abilities generalize to downstream evaluation on visual math questions (Tab.~\ref{tab:ablation_prompt}). \textbf{Bold text} indicates reasoning instructions synthesized by GPT-4o~\citep{hurst2024gpt}.

\label{sec:prompt}

\hypertarget{A3}{}\subsection{Detail of Format Reward}
\label{sec:detail_reward}
\enlargethispage*{0pt}
The format reward $r_\text{format}$ validates whether the response follows the task-specific format:
\begin{equation}
    r_\text{format} = 
    \begin{cases} 
    0.1, & \text{if the response follows the required format} \\
    0, & \text{otherwise} 
    \end{cases}
\end{equation}
For Snake game, the desired format is:
\begin{align*}
\small
\texttt{<think>}...\texttt{</think>}\texttt{<best\_answer>}...\texttt{</best\_answer>}\texttt{<worst\_answer>}...\texttt{</worst\_answer>}\,.
\end{align*}
As suggested by the format, we encourage the model to predict both a positive move that moves toward the apple and a negative move that leads to failure. This reward encourages contrastive decision-making, which not only improves the model’s gameplay abilities but also boosts downstream reasoning performance on visual math benchmarks. We ablate the effect in Tab.~\ref{tab:ablation_reward}.
For the rotation task, the required format is 
simply $\small \texttt{<think>}...\texttt{</think>}\texttt{<answer>}...\texttt{</answer>}.$

\section{Evaluation}

\hypertarget{B1}{}\subsection{Evaluation Detail of Atari Game}
\label{sec:atarigame_appendix}
\enlargethispage*{0pt}
\begin{figure}[H]
    \centering
    \begin{subfigure}{\textwidth}
        \centering
        \includegraphics[height=4.7cm]{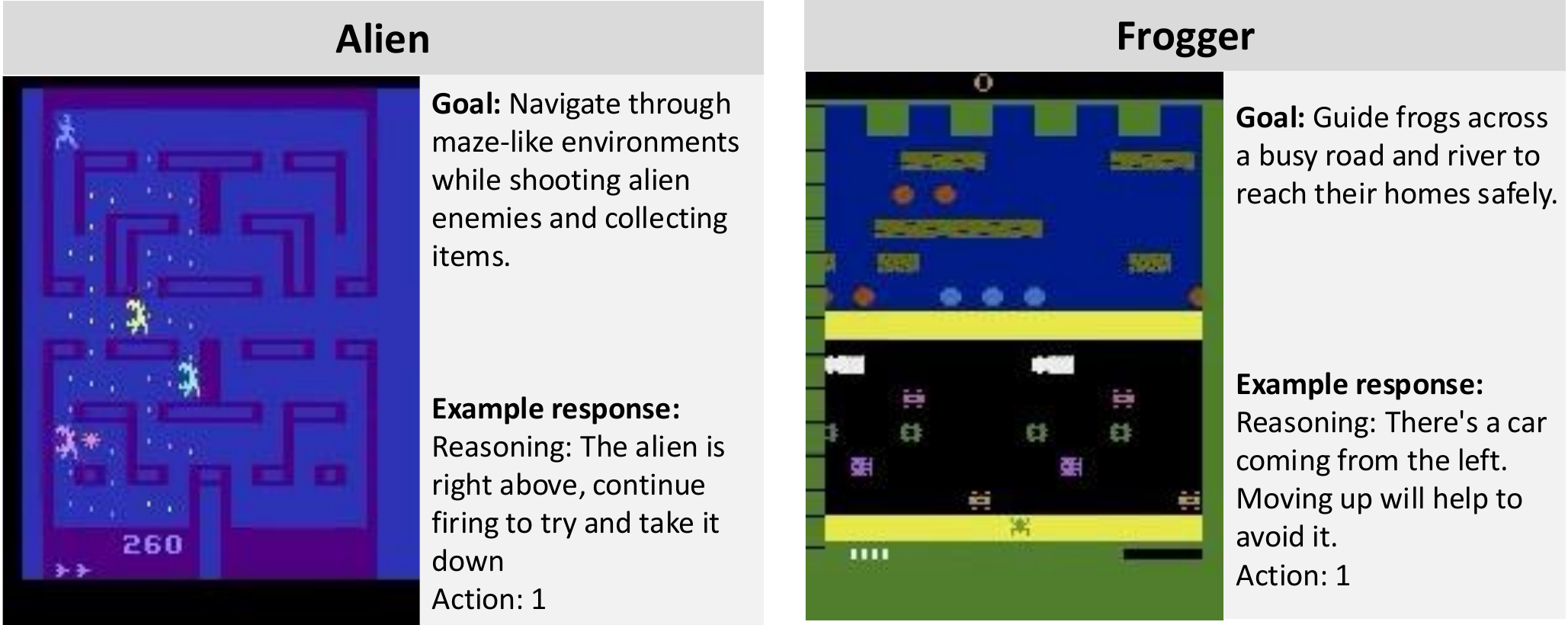}
    \end{subfigure} 
    \begin{subfigure}{\textwidth}
        \centering
        \includegraphics[height=4.7cm]{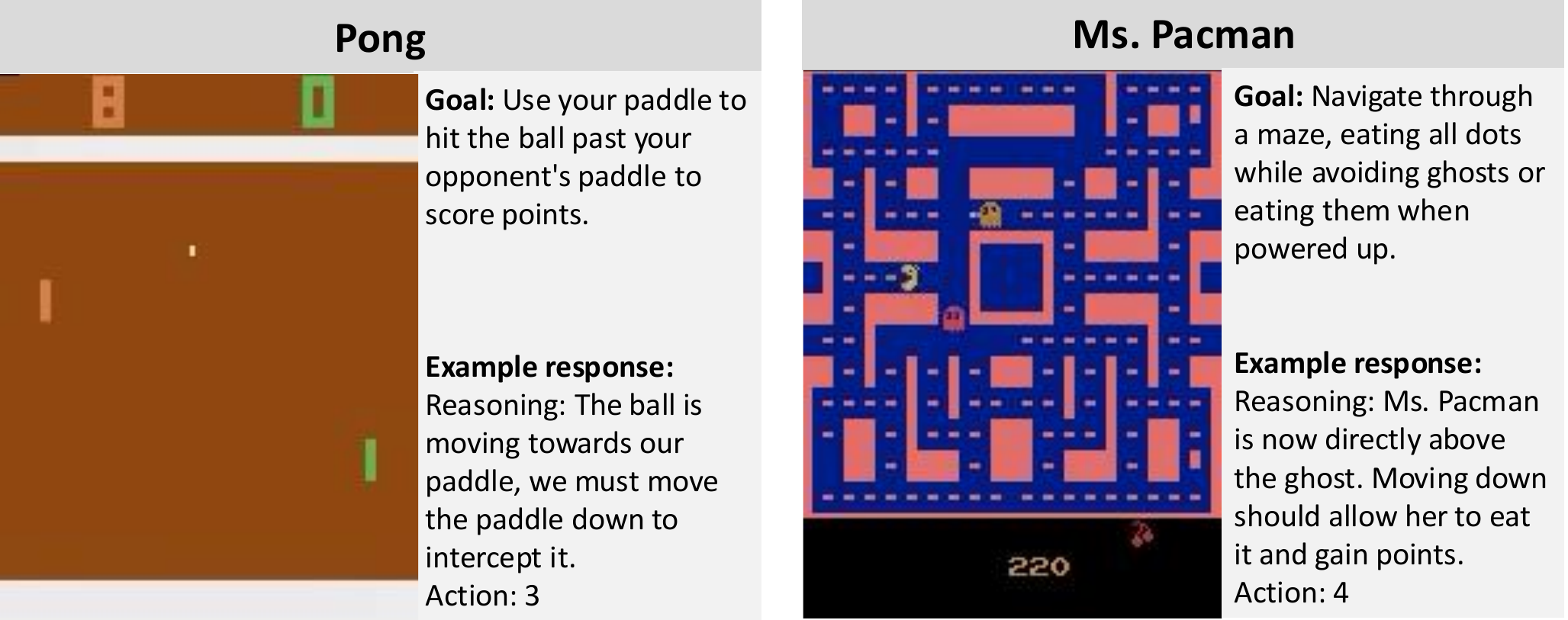}
    \end{subfigure}
    \begin{subfigure}{\textwidth}
        \centering
        \includegraphics[height=4.7cm]{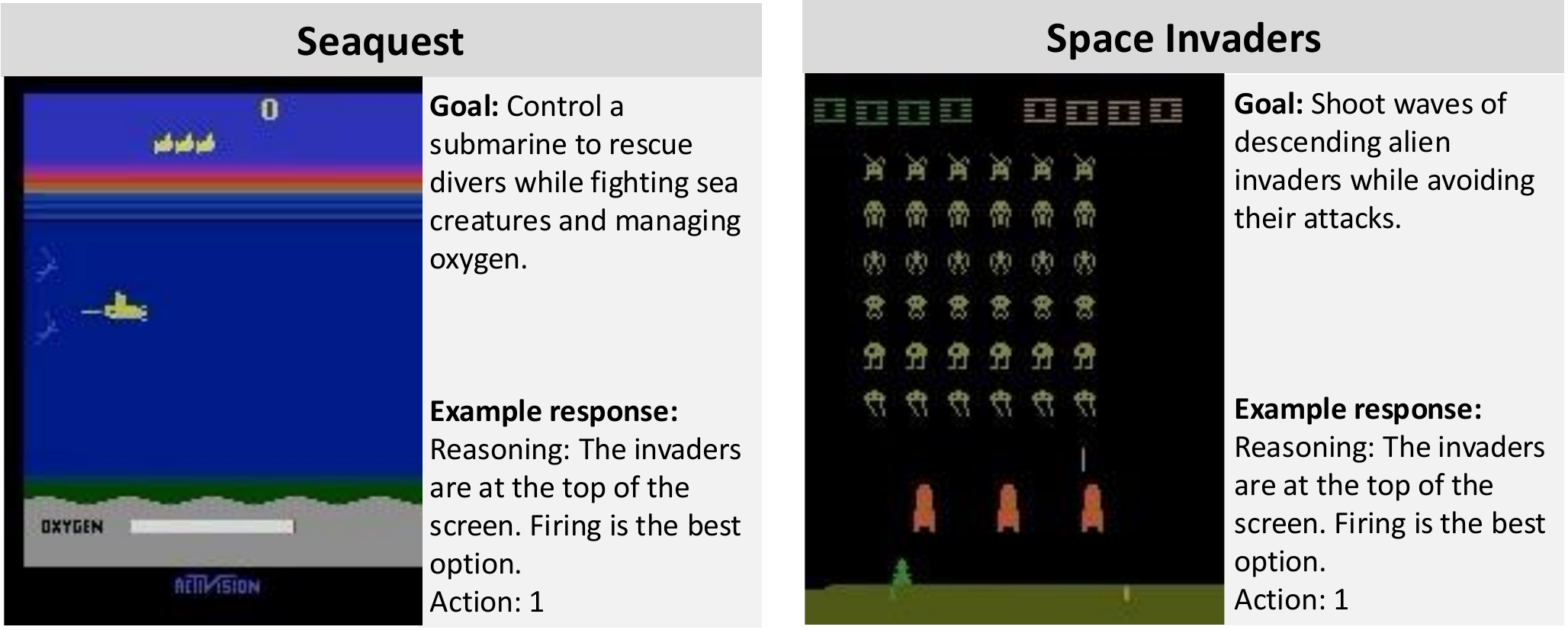}
    \end{subfigure}  
    \begin{subfigure}{\textwidth}
        \centering
        \includegraphics[height=4.7cm]{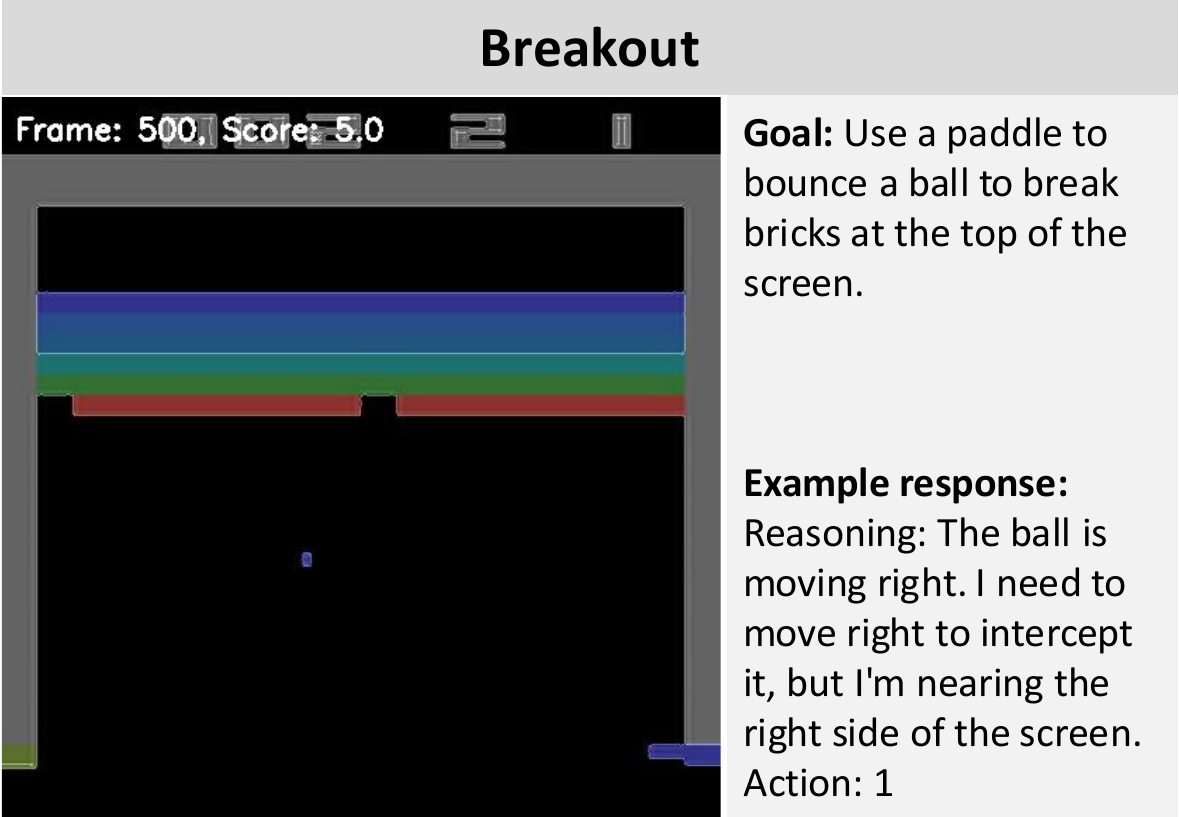}
    \end{subfigure}
\caption{\textbf{Goal and example response from model of Atari games used for evaluation.} We implement 7 kinds of Atari games from Atari-GPT~\citep{waytowich2024atari}.}
    \label{fig:atarigame}
\end{figure}

To evaluate out-of-distribution generalization, we test \ourmethod on Atari-GPT~\citep{waytowich2024atari}, a benchmark for evaluating MLLMs as decision-making agents in Atari video games, as shown in Fig.~\ref{fig:atarigame}. The benchmark consists of seven different Atari games: Alien, Frogger, Pong, Ms. Pacman, Seaquest, Space Invaders and Breakout. These games present diverse visual environments which is different from Snake game and Rotation game, and require different strategic approaches to finish the goal, making them an ideal test bed for \ourmethod evaluating out-of-distribution generalization capabilities.

For evaluation, we input game frames as pixel observations to our model, following the established protocol in Atari-GPT. Specifically, each game frame is resized from $210 \times 160 \times 3$ to $512 \times 512 \times 3$, then provided to our model along with game-specific action information. We maintain a context buffer containing the two previous frames and responses together with the current frame to enable temporal reasoning. Following Atari-GPT, we implement frame skipping of 8 frames, which extends the standard 4-frame skipping in ALE to reduce computational intensity while preserving gameplay continuity.

We evaluate our method through four independent rollouts of 1,000 timesteps each and report the average cumulative reward, with results presented in Tab.~\ref{tab:atarigame}.

\label{sec:atarigame}

\hypertarget{B2}{}\subsection{Ablation On Rotation Game}
\begin{table}[H]
\centering
\caption{\textbf{Ablation study.} Similar to the evaluation in Tab.~\ref{tab:ablation}, we analyze how different aspects of our post-training strategy within the Rotation game affect downstream generalization benchmarks. The base model is Qwen2.5-VL-7B, with results shown in \textcolor[gray]{0.5}{gray}. The default settings from Tab.~\ref{tab:math_generalization} and Tab.~\ref{tab:reason_generalization} are highlighted in \colorbox[HTML]{CFEFFF}{blue}. We observe the same improvement trends for each strategy as reported in Tab.~\ref{tab:ablation}.}
\begin{subtable}[!t]{0.48\textwidth}
\centering
\tablestyle{4.5pt}{1.2}
\caption{\centering Prompt design.}
\vspace{-5pt}
\label{tab:ablation_prompt_rot}
\resizebox{\linewidth}{!}{
\begin{tabular}{y{100}cccc}
prompt & \textbf{Avg.} & \textbf{Math} & \textbf{CLEVR+} & \textbf{Geo.} \\ 
\midrule 
\textcolor[gray]{0.5}{base model} & \textcolor{gray}{49.1} & \textcolor{gray}{47.7} & \textcolor{gray}{54.9} & \textcolor{gray}{44.8} \\ 
w/o \prompt & 61.4 & 48.9 & 80.4 & 54.8 \\ 
\mycolor{w/ \prompt} & \mycolor{62.6} & \mycolor{49.3} & \mycolor{80.7} & \mycolor{57.9} \\ 
\end{tabular}
}
\end{subtable}
\hfill
\begin{subtable}[!t]{0.48\textwidth}
\centering
\tablestyle{4.5pt}{1.2}
\caption{\centering SFT vs. RL.}
\vspace{-5pt}
\label{tab:ablation_sftvsrl_rot}
\resizebox{\linewidth}{!}{
\begin{tabular}{y{100}cccc}
post-training & \textbf{Avg.} & \textbf{Math} & \textbf{CLEVR+} & \textbf{Geo.} \\ 
\midrule 
\textcolor[gray]{0.5}{base model} & \textcolor{gray}{49.1} & \textcolor{gray}{47.7} & \textcolor{gray}{54.9} & \textcolor{gray}{44.8} \\ 
SFT & 55.6 & 44.0 & 75.4 & 47.5 \\
\mycolor{RL} & \mycolor{62.6} & \mycolor{49.3} & \mycolor{80.7} & \mycolor{57.9} \\ 
\end{tabular}
}
\end{subtable}
\hfill
\begin{subtable}[H]{0.48\textwidth}
\centering
\tablestyle{4.5pt}{1.2}
\caption{\centering Difficulty control.}
\vspace{-5pt}
\label{tab:ablation_difficulty_rot}
\resizebox{\linewidth}{!}{
\begin{tabular}{y{100}cccc}
difficulty control & \textbf{Avg.} & \textbf{Math} & \textbf{CLEVR+} & \textbf{Geo.} \\ 
\midrule 
\textcolor[gray]{0.5}{base model} & \textcolor{gray}{49.1} & \textcolor{gray}{47.7} & \textcolor{gray}{54.9} & \textcolor{gray}{44.8} \\ 
w/o  difficulty control & 61.0 & 48.0 & 80.2 & 54.8 \\ 
\mycolor{w/ difficulty control} & \mycolor{62.6} & \mycolor{49.3} & \mycolor{80.7} & \mycolor{57.9} \\ 
\end{tabular}
}
\end{subtable}
\label{tab:ablation_rot}
\end{table}
\label{sec:ablation_rot}

As shown in Tab.~\ref{tab:ablation_rot}, we conduct a similar ablation study to Tab.~\ref{tab:ablation}, but replace the Snake game environment with the Rotation game. Our results demonstrate the same consistent improvement trends on downstream generalization benchmarks for each strategy employed. 

Specifically, we control the task difficulty by varying the rotation angles between two images. In the uncontrolled difficulty setting, the rotation angle between images can be clockwise 90°, counter-clockwise 90°, or 180°. However, we found that explicitly requiring the model to distinguish between clockwise and counter-clockwise rotations leads to training difficulties. Therefore, we remove it and only retain option of clockwise 90° and 180° rotations.
\enlargethispage*{0pt}

Unlike the Snake game, we cannot conduct the ablations shown in Tab.~\ref{tab:ablation_vision} because the Rotation game is inherently vision-dependent and requires visual input. Similarly, we cannot perform the ablations in Tab.~\ref{tab:ablation_reward} because the Rotation game provides only binary answer options, making it impossible to meaningfully designate both "best" and "worst" answers simultaneously.
\begin{figure}[H]
    \centering
    \includegraphics[width=0.8\linewidth]{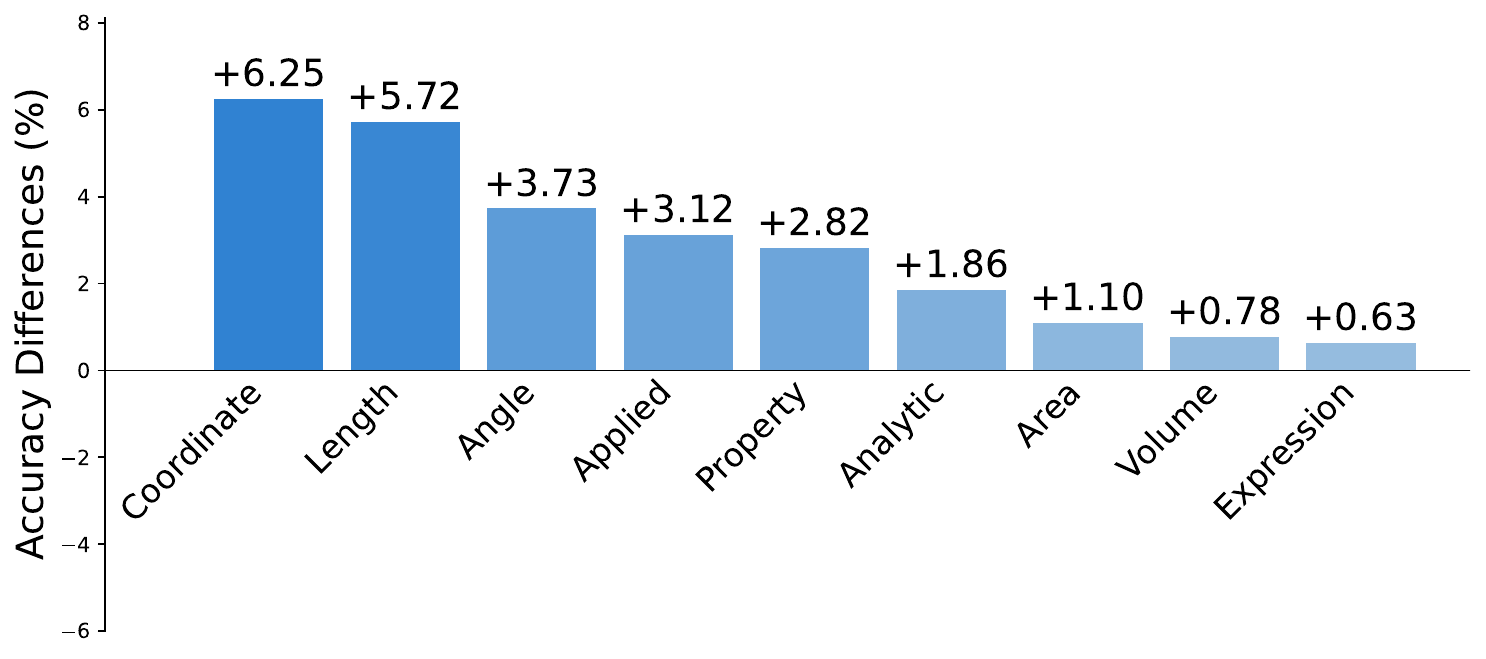}
    \caption{\textbf{Accuracy differences between \ourmethod-Snake+Rotation and base model without RL training across mathematical subfields in Mathverse.} The synergistic effects of jointly training on two games observed suggest that complementary games can enhance overall mathematical reasoning capabilities.}
    \label{fig:acc_diff_appendix}
\end{figure}
\hypertarget{B4}{}\subsection{Synergistic Effects of Multi-Game Training}
\label{sec:acc_diff_appendix}
As discussed in Sec.~\ref{sec:acc_diff}, our analysis reveals that each game develops distinct reasoning abilities in the model. 
To investigate potential combined benefits, we conducted experiments where models were trained simultaneously on \textit{both} the Snake and Rotation games.
Fig.~\ref{fig:acc_diff_appendix} shows that joint training effectively combines the strengths of each individual game, improving performance across the mathematical areas where each game shows particular effectiveness, resulting in greater overall gains on Mathverse.
These results suggest that strategically combining games with complementary strengths offers a simple yet effective approach to enhance model generalization abilities.

\hypertarget{B4}{}\subsection{Reasoning Ability Boundary via Pass$@k$ Evaluation}
\begin{figure}[H]
    \centering
    \includegraphics[width=0.75\linewidth]{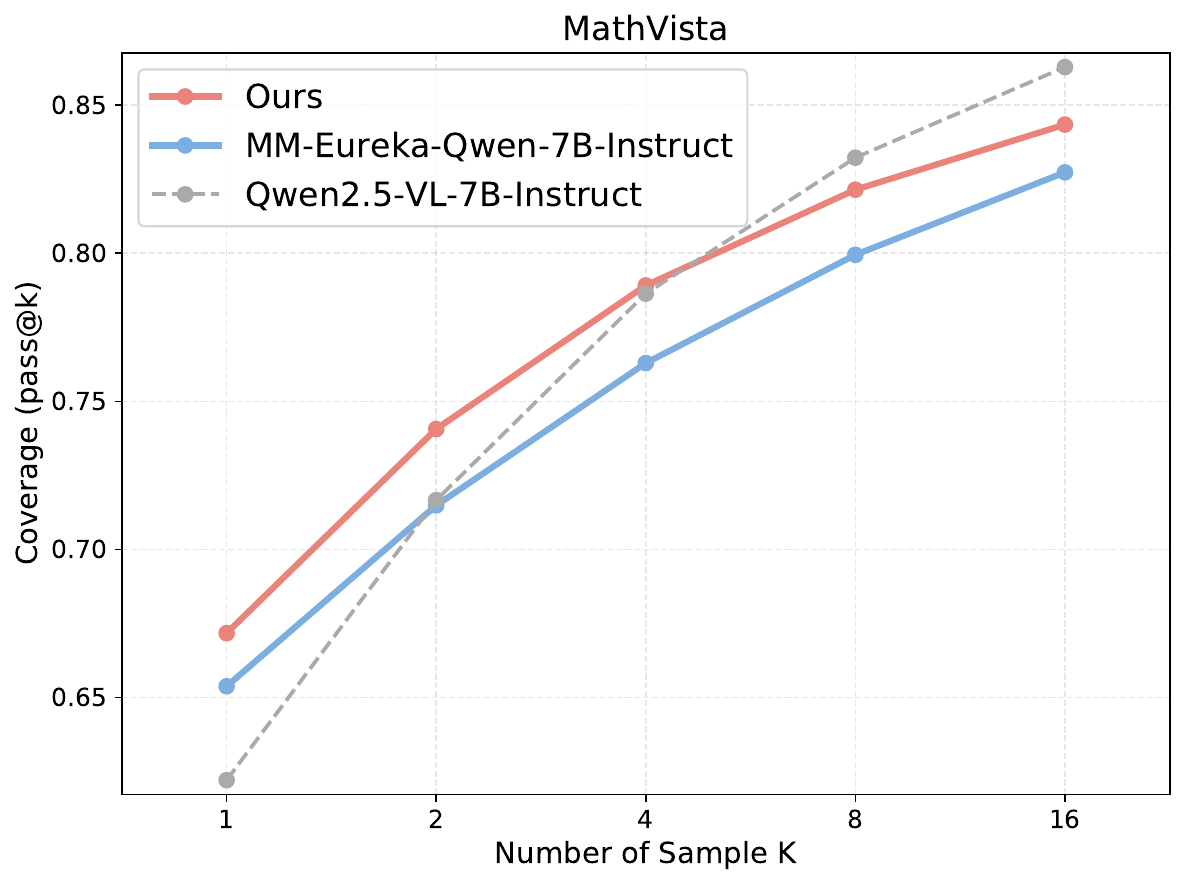}
\caption{Pass$@k$ performance curves on MathVista comparing base models with their zero-RL counterparts trained on mathematical data and game data, respectively.}
    \label{fig:passk}
\end{figure}
We explore the reasoning ability boundary of models trained with different RL approaches by evaluating the pass$@k$ metric. This metric measures the probability that at least one of $k$ independent model samples solves a given problem, indicating the true scope or boundary of a model's reasoning capability - essentially what problems the model can potentially solve given enough sampling attempts.

We evaluate the pass$@k$ performance of three models: the Base Model without RL training, MM-Eureka-Qwen-7B-Instruct, and our \ourmethod. As shown in Fig.~\ref{fig:passk}, our \ourmethod consistently demonstrates increasing pass$@k$ scores on Mathverse as $k$ increases. This finding suggests that our approach can effectively solve complex problems when allowed multiple reasoning attempts, uncovering capabilities not apparent in single-sample evaluations.

Moreover, compared to the other RL-trained model, MM-Eureka-Qwen-7B-Instruct, our model achieves a steeper improvement in pass$@k$ as $k$ increases. This indicates that \ourmethod possesses a broader reasoning boundary and stronger reasoning abilities, enabling it to solve a wider range of problems when given sufficient opportunities to explore different solution paths. 

Finally, our results demonstrate that as $k$ increases, base models without RL training eventually outperform RL-trained models. This aligns with the findings in \citep{yue2025does} that highlight a fundamental limitation of reinforcement learning with verifiable rewards (RLVR): while RL training significantly improves performance at small $k$ values (e.g., pass$@1$), base models possess a wider coverage of solvable problems. This suggests a trade-off where RL optimization focuses on solving high-probability problems at the expense of broader solution coverage. Future work should explore RLVR algorithms that can improve pass$@k$ performance across all values of $k$, effectively extending the reasoning boundary beyond that of the base model.

\hypertarget{B5}{}\subsection{Detail of Evaluation Benchmarks}
\label{sec:detailed_eval_benchmarks}
\begin{figure}[t]
  \centering
\begin{subfigure}[t]{0.19\textwidth}
    \centering
    \includegraphics[width=\linewidth]{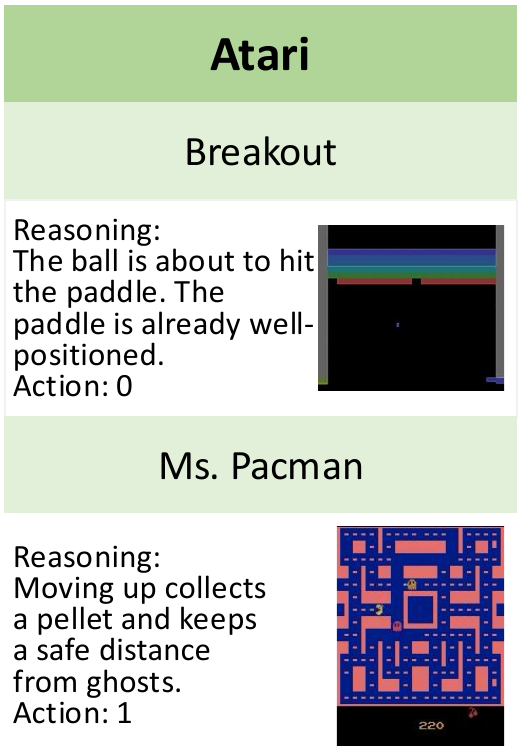}
    \caption{\centering Atari games.}
    \label{fig:dataset_card_1}
\end{subfigure}%
\begin{subfigure}[t]{0.78\textwidth}
    \centering
    \includegraphics[width=\linewidth]{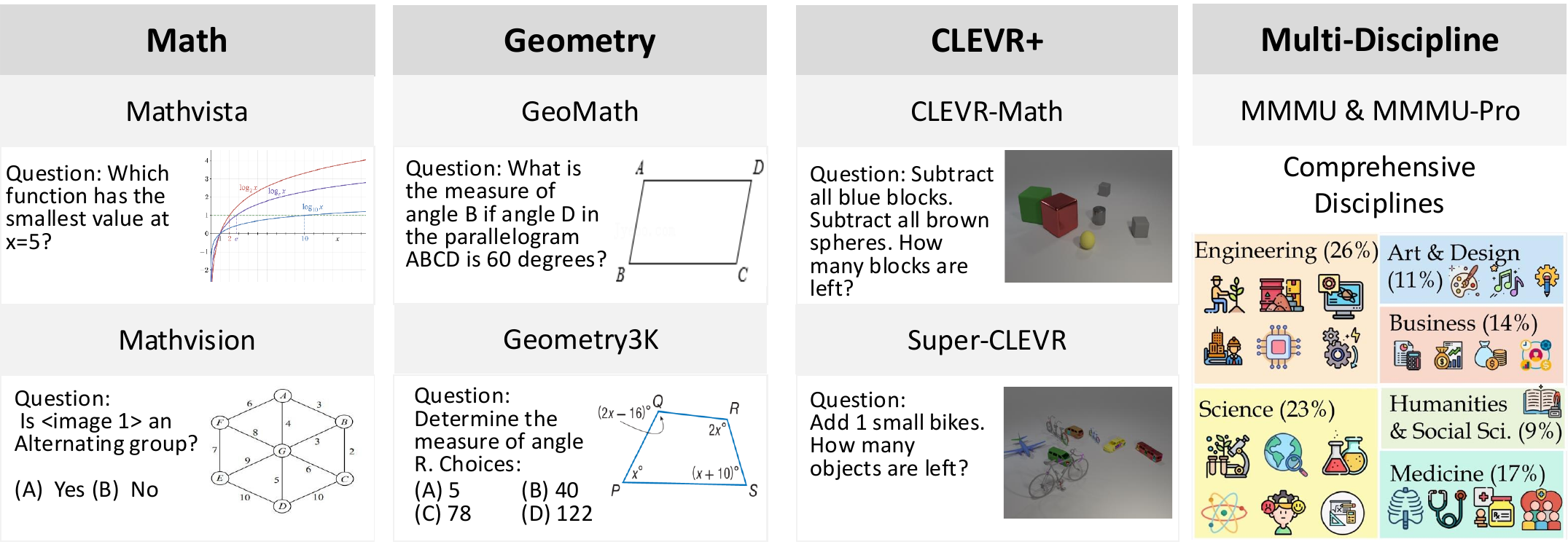}
    \caption{\centering Out-of-domain tasks.}    
    \label{fig:dataset_card_2}
\end{subfigure}%
  \caption{\textbf{Samples from our generalization reasoning benchmarks.} We evaluate the proposed \ourmethod with two types of generalization: (a) \textit{out-of-distribution} generalization, where models trained on our visual games are tested on unseen Atari games~\citep{waytowich2024atari}; and (b) \textit{out-of-domain} generalization, where models trained only on game tasks are evaluated on diverse multimodal reasoning tasks including mathematical reasoning, geometric problem-solving, 3D understanding on CLEVR+ and multi-discipline reasoning on MMMU series.}
  \label{fig:dataset}
  \vspace{-5pt}
\end{figure}
To obtain a clearer picture of the various facets of MLLM performance, we follow prior studies~\citep{tong2024cambrian, li2024survey} and systematically and carefully divide existing benchmarks into two broad groups:  
(i) \emph{reasoning-oriented benchmarks}, which require multi-step or mathematical reasoning to solve the problems, and  
(ii) \emph{general-purpose perception benchmarks}, which primarily assess broad visual understanding and perception abilities.

For reasoning-oriented benchmarks, we comprehensively evaluate the visual reasoning generalization capabilities of RL through gaming on a diverse collection of tasks that specifically demand advanced visual reasoning skills, including math-focused tasks like Math and Geometry, and other comprehensive reasoning benchmarks beyond math, like CLEVR+ and Multi-Discipline. Fig.~\ref{fig:dataset_card_2} illustrates specific examples from each benchmark.

\begin{itemize}
    \item \textbf{Math} evaluates multimodal math reasoning with widely-used datasets: MathVista (testmini)~\citep{lu2024mathvistaevaluatingmathematicalreasoning}, MathVerse (testmini)~\citep{zhang2024mathversedoesmultimodalllm}, and MathVision (test)~\citep{wang2024measuring}. MathVista offers diverse problems spanning VQA, logic, algebra, and geometry; MathVerse emphasizes algebraic and geometric image comprehension; MathVision tests abstract visual reasoning.
    
    \item \textbf{Geometry} evaluates structural interpretation skills across mathematical diagrams, medical images, charts, and architectural layouts. It uses datasets GeoMath (Geo170K~\citep{geo170k}, Math360K~\citep{shi2024mathllavabootstrappingmathematicalreasoning}) and Geometry3K~\citep{lu2021inter}, featuring both choice and non-choice questions. Following Reason-RFT~\citep{tan2025reason}, we test with 820 GeoMath and 800 Geometry3K samples.
    
    \item \textbf{CLEVR+} evaluates the integration of mathematical and spatial reasoning skills through challenging arithmetic problems in complex 3D block-based scenes, including sub-tasks on CLEVR-Math~\citep{clevr-math} and Super-CLEVR~\citep{super-clevr}. Following Reason-RFT~\citep{tan2025reason}, we use 1K test samples from each of CLEVR-Math and Super-CLEVR.
    
    \item \textbf{Multi-Discipline} evaluates college-level expert knowledge across six disciplines: Art \& Design, Business, Science, Health \& Medicine, Humanities \& Social Science, and Tech \& Engineering. We follow the evaluation setting of MMMU~\citep{yue2024mmmu} val set (900 questions) and MMMU-Pro~\citep{yue2024mmmupro} overall score (average of standard 10-option and vision-only settings).
\end{itemize}

For general-purpose perception benchmarks, we systematically evaluate comprehensive visual capabilities. Following previous work, these benchmarks are categorized into three distinct types: General, Vision-Centric, and OCR \& Chart.

\begin{itemize}
    \item \textbf{General} benchmarks assess fundamental visual understanding capabilities. We evaluate MuirBench~\citep{wang2024muirbench} for multi-image understanding and CRPE~\citep{kazemzadeh2014referitgame} for relation understanding.
    
    \item \textbf{Vision-Centric} benchmarks thoroughly evaluate perception, real-world understanding, and multi-modal capabilities. We assess MMVP~\citep{tong2024eyes}, RealWorldQA~\citep{grok15}, MMStar~\citep{chen2024we}, MME~\citep{fu2023mme}, and BLINK~\citep{fu2024blink}.
    
    \item \textbf{OCR \& Chart} understanding benchmarks focus on text-rich visual content. We specifically use AI2D~\citep{kembhavi2016diagram} for diagram understanding, SEED-Bench-2-Plus~\citep{li2024seed2plus} for text-rich visual comprehension, DocVQA~\citep{docvqa} for document understanding, and OCRBench~\citep{liu2024ocrbenchhiddenmysteryocr} for comprehensive OCR evaluation.
\end{itemize}

For 3D spatial reasoning, we evaluate on VSI-Bench~\citep{yang2025thinking}. We run the benchmark through the LMMs-Eval~\citep{zhang2025lmms} harness with the standard VSI Bench configuration, use greedy decoding for inference, and uniformly sample 32 frames per video across the full duration. 
\hypertarget{B6}{}\subsection{Inference Length Analysis}

Recent reinforcement learning studies~\citep{xie2025logic,aggarwal2025l1} have raised questions about whether performance improvements stem from genuinely enhanced reasoning capabilities or merely from models generating longer responses. To address this concern, we analyze the relationship between response length and performance for models trained with our game-based approach.

\begin{table}[h]
\centering
\begin{tabular}{lcc}
\hline
\textbf{Model} & \textbf{Response Length} & \textbf{Math Avg.} \\
\hline
Qwen2.5-VL-7B (baseline) & 250 & 47.7 \\
ViGaL (ours, RL on games) & 268 & \textbf{50.6} \\
\hline
\end{tabular}
\caption{Response length and performance on visual math benchmarks. Our game-based RL approach achieves significant performance gains while maintaining comparable inference costs.}
\label{tab:response_length}
\end{table}

Table~\ref{tab:response_length} demonstrates that our performance improvements are not simply due to increased verbosity. Our ViGaL model achieves substantial performance gains (50.6\% vs. 47.7\%) while maintaining nearly identical inference costs—the response length increases by only 7\% (268 vs. 250 tokens). This minimal increase in response length, coupled with the significant accuracy improvement, indicates that the model has learned transferable skills rather than merely generating longer outputs.

These results suggest that game-based RL training enables effective knowledge transfer from game environments to mathematical problem-solving. For example, spatial reasoning skills acquired from the Rotation game and coordinate recognition abilities developed through Snake gameplay transfer effectively to visual math tasks. The model thus learns genuine problem-solving strategies while maintaining inference costs.
\hypertarget{B7}{}\subsection{Reasoning Correlation Analysis Between Game and Math}
\label{sec:game_math_rel}
To understand the mathematical reasoning patterns in snake game playing, we developed a systematic approach to extract and analyze reasoning steps from multiple gameplay traces. Our methodology uses GPT-5~\cite{2025gpt5} as an analytical tool in a two-stage process. 

In Stage A, we collect multiple snake game "thinking traces", which are detailed reasoning sequences generated during gameplay, and distill them into a generalized set of 8 core reasoning steps. These steps abstract away specific details like exact coordinates or particular board configurations to capture fundamental cognitive operations. The operations include parsing board state, enumerating moves, safety screening, path metric selection, distance computation, target identification, enemy anticipation, and move ranking. This summarization ensures our analysis focuses on transferable reasoning patterns rather than game-specific instances.

In Stage B, we quantify how mathematical each reasoning step is by evaluating its correlation with nine distinct mathematical aspects. We use a simple 3-level scoring system where 0 means no correlation, 1 means low correlation, and 2 means high correlation. GPT-5 analyzes how strongly each step relates to mathematical concepts such as coordinate manipulation, distance metrics, analytical reasoning, and geometric properties. 

The resulting correlation matrix in Tab.~\ref{tab:game_math_rel} reveals clear patterns. Coordinate-based reasoning dominates steps that involve spatial parsing and movement planning, particularly Steps 1 through 3, Step 5, and Step 7. Meanwhile, analytical and length-based reasoning become prominent in optimization steps like target identification and move ranking, seen in Steps 6 and 8. Steps 4 and 5, which involve path metrics and distance computation, show high correlation with both coordinate systems and length calculations. This confirms the geometric nature of pathfinding in grid-based environments. Our systematic analysis demonstrates that even seemingly simple game-playing behaviors require sophisticated integration of multiple mathematical reasoning capabilities.

\begin{tcolorbox}[
    title={Prompt Template for Reasoning Step Extraction and Correlation Analysis},
    colback=blue!5!white,
    colframe=blue!40!white,
    left=1mm, right=1mm, top=1mm, bottom=1mm,
    width=\textwidth,
    center,
    fonttitle=\small,
    fontupper=\small,
    label=prompt:correlation
]
\textbf{Stage A - Step Extraction:}\\
Given multiple snake game thinking traces, extract N general reasoning steps (6-9 steps) that capture the core operations. Abstract away instance-specific details and output:
\begin{itemize}
    \item Short, action-oriented step names with one-line descriptions
    \item General patterns covering: state parsing, move generation, safety screening, target selection via distance, opponent awareness, scoring/tie-breaks, decision, reporting
\end{itemize}

\textbf{Stage B - Mathematical Aspect Correlation:}\\
For each extracted step, assign correlation levels (0/1/2) to these mathematical aspects:
\begin{itemize}
    \item \textbf{Expression}: Formatting/structuring outputs
    \item \textbf{Coordinate}: Reading/writing positions, mapping moves to (x,y)
    \item \textbf{Area}: Board regions/bounds as areas
    \item \textbf{Volume}: 3D spatial reasoning (if applicable)
    \item \textbf{Applied}: Goal-directed task execution
    \item \textbf{Property}: Rules/invariants (bounds, occupancy, collision)
    \item \textbf{Angle}: Angle-based path reasoning
    \item \textbf{Analytic}: Selection/optimization, tie-break logic
    \item \textbf{Length}: Distance metrics (Manhattan/L1, grid paths)
\end{itemize}
Output as structured table with integer scores only (0 = no correlation, 1 = low, 2 = high).
\end{tcolorbox}.

\definecolor{nocorr}{RGB}{255,255,255}      
\definecolor{lowcorr}{RGB}{200,230,255}     
\definecolor{highcorr}{RGB}{50,100,200}     

\begin{table}[htbp]
\centering
\caption{Correlation Matrix of each step reasoning trace of playing snake game with solving math questions. (0=No Correlation, 1=Low Correlation, 2=High Correlation)}
\label{tab:game_math_rel}
\resizebox{\textwidth}{!}{%
\begin{tabular}{>{\centering\arraybackslash}p{0.8cm} p{5cm} *{9}{>{\centering\arraybackslash}p{0.8cm}}}
\toprule
\textbf{Step} & \textbf{Operation} & \rotatebox{90}{\textbf{Expression}} & \rotatebox{90}{\textbf{Coordinate}} & \rotatebox{90}{\textbf{Area}} & \rotatebox{90}{\textbf{Volume}} & \rotatebox{90}{\textbf{Applied}} & \rotatebox{90}{\textbf{Property}} & \rotatebox{90}{\textbf{Angle}} & \rotatebox{90}{\textbf{Analytic}} & \rotatebox{90}{\textbf{Length}} \\
\midrule
1 & Parse the board state & \cellcolor{lowcorr} & \cellcolor{highcorr} & \cellcolor{lowcorr} & \cellcolor{nocorr} & \cellcolor{lowcorr} & \cellcolor{lowcorr} & \cellcolor{nocorr} & \cellcolor{lowcorr} & \cellcolor{nocorr} \\
2 & Enumerate candidate moves & \cellcolor{lowcorr} & \cellcolor{highcorr} & \cellcolor{nocorr} & \cellcolor{nocorr} & \cellcolor{lowcorr} & \cellcolor{lowcorr} & \cellcolor{nocorr} & \cellcolor{lowcorr} & \cellcolor{nocorr} \\
3 & Safety screening (worst-move test) & \cellcolor{lowcorr} & \cellcolor{highcorr} & \cellcolor{lowcorr} & \cellcolor{nocorr} & \cellcolor{highcorr} & \cellcolor{highcorr} & \cellcolor{nocorr} & \cellcolor{lowcorr} & \cellcolor{nocorr} \\
4 & Choose a path metric & \cellcolor{highcorr} & \cellcolor{highcorr} & \cellcolor{nocorr} & \cellcolor{nocorr} & \cellcolor{lowcorr} & \cellcolor{lowcorr} & \cellcolor{lowcorr} & \cellcolor{lowcorr} & \cellcolor{highcorr} \\
5 & Compute distances to apples & \cellcolor{highcorr} & \cellcolor{highcorr} & \cellcolor{nocorr} & \cellcolor{nocorr} & \cellcolor{lowcorr} & \cellcolor{lowcorr} & \cellcolor{lowcorr} & \cellcolor{lowcorr} & \cellcolor{highcorr} \\
6 & Identify nearest target & \cellcolor{lowcorr} & \cellcolor{lowcorr} & \cellcolor{nocorr} & \cellcolor{nocorr} & \cellcolor{highcorr} & \cellcolor{lowcorr} & \cellcolor{nocorr} & \cellcolor{highcorr} & \cellcolor{highcorr} \\
7 & Anticipate enemy motion (when present) & \cellcolor{lowcorr} & \cellcolor{highcorr} & \cellcolor{lowcorr} & \cellcolor{nocorr} & \cellcolor{lowcorr} & \cellcolor{highcorr} & \cellcolor{lowcorr} & \cellcolor{highcorr} & \cellcolor{nocorr} \\
8 & Rank safe moves & \cellcolor{lowcorr} & \cellcolor{lowcorr} & \cellcolor{nocorr} & \cellcolor{nocorr} & \cellcolor{highcorr} & \cellcolor{lowcorr} & \cellcolor{lowcorr} & \cellcolor{highcorr} & \cellcolor{highcorr} \\
\bottomrule
\end{tabular}
}
\vspace{0.3cm}\\
\textbf{Correlation Legend:}\\
\vspace{0.1cm}
\colorbox{nocorr}{\phantom{\textbf{XX}}} = No Correlation \quad
\colorbox{lowcorr}{\phantom{\textbf{XX}}} = Low Correlation \quad
\colorbox{highcorr}{\phantom{\textbf{XX}}} = High Correlation
\end{table}

\hypertarget{C}{}\section{Case Study}
\label{sec:case_study}

\begin{figure}[H]
    \centering
    \begin{subfigure}{\linewidth}
        \centering
        \includegraphics[width=0.8\linewidth]{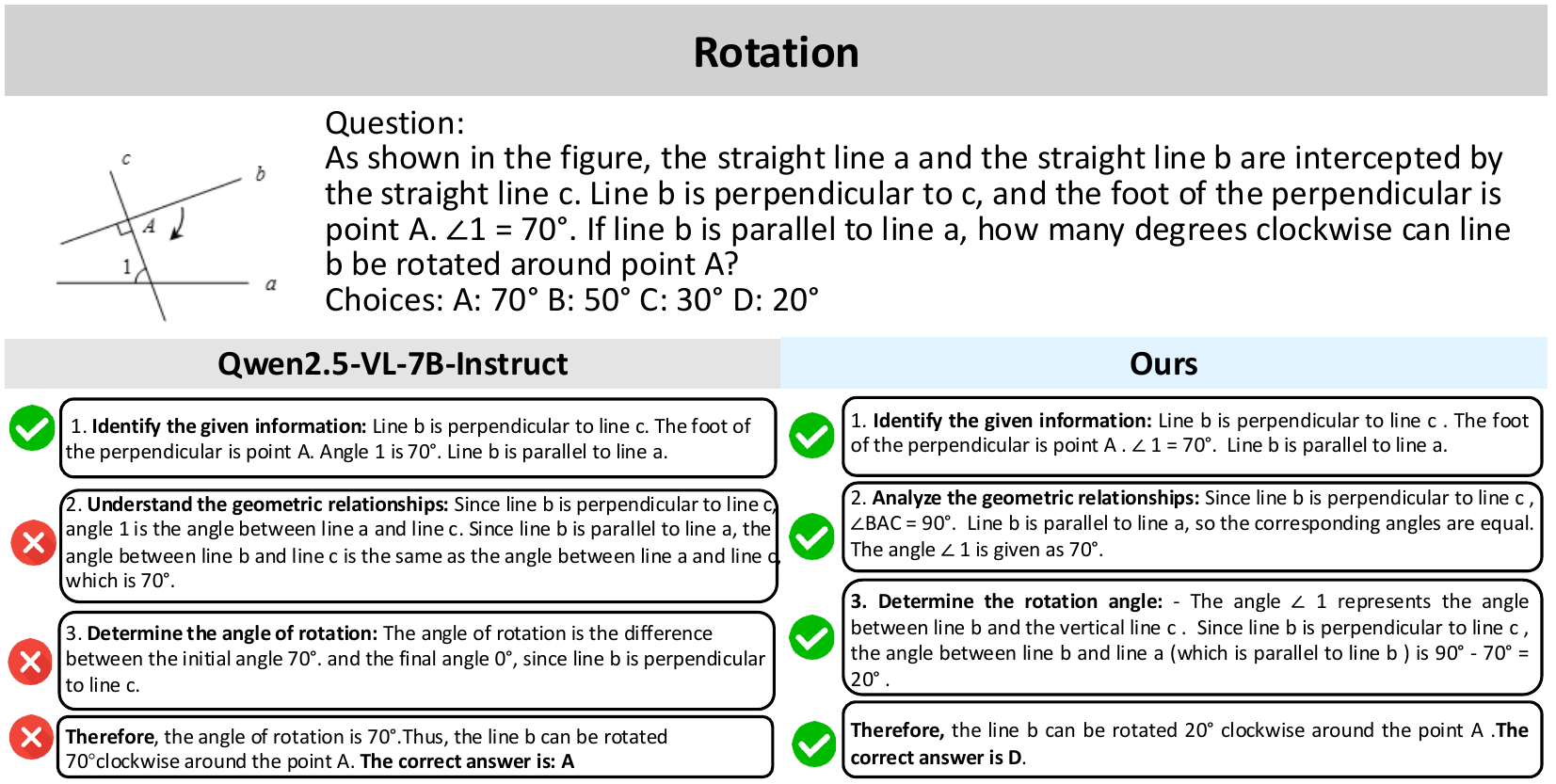}
        \caption{A case study from Mathverse. Base model misinterpreted the geometric configuration and rotation direction, while our model correctly identified the perpendicular relationship and calculated the proper angle.}
        \label{fig:case_study_1}
    \end{subfigure}
    
    \vspace{1em}
    
    \begin{subfigure}{\linewidth}
        \centering
        \includegraphics[width=0.8\linewidth]{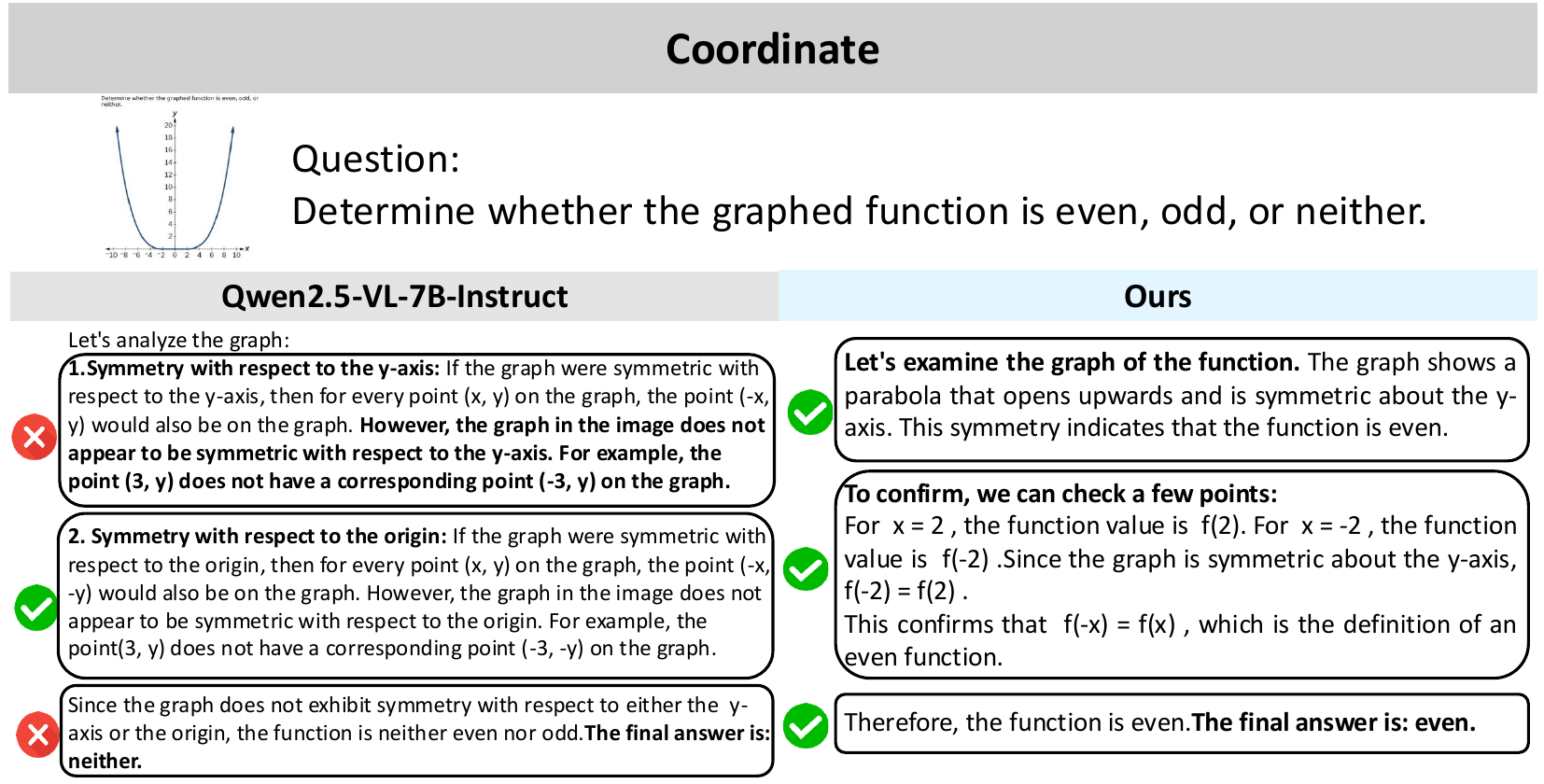}
        \caption{A case study from Mathverse. Base model misperceived critical visual information like symmetry and coordinates in graphs, while our model demonstrated accurate visual perception for mathematical elements.}
        \label{fig:case_study_2}
    \end{subfigure}
    \caption{Comparison of base model and our model after rule-based RL training, showing improved visual-mathematical reasoning on geometric and coordinate problems.}
    \label{fig:case_study}
\end{figure}

We provide quantitative comparison examples below to demonstrate reasoning improvements on mathematical problems after RL training. 
In Fig.~\ref{fig:case_study_1}, when solving a geometric angle problem, the base model fails to correctly interpret the critical relationship between perpendicular lines and corresponding angles. It makes contradictory assumptions about angle measures, leading to an incorrect calculation of the required rotation. In contrast, our \ourmethod precisely tracks the geometric constraints and properly calculates the angle difference between initial and target positions. In Fig.~\ref{fig:case_study_2}, when analyzing function properties from a graph, the base model incorrectly claims the function lacks symmetry despite clear visual evidence. It fails to recognize the fundamental y-axis symmetry of the parabola shown in the image. Our model immediately identifies this critical symmetrical pattern and correctly applies the appropriate mathematical definition of an even function, demonstrating enhanced visual perception of mathematical structures.

\end{document}